\documentclass{article} 
\newcommand{\projecturl}{https://kaustubhp11.github.io/querylm/}
\newcommand{\projectlabel}{Project Page}
\newcommand{\hfurl}{https://huggingface.co/collections/Kausp11/search-native-pretraining}
\newcommand{\codeurl}{https://github.com/KaustubhP11/QueryLM}
\newif\ifshowcodelink\showcodelinktrue 
\usepackage{iclr2027_conference,times}
\usepackage[table]{xcolor}

\usepackage{amsmath,amsfonts,bm}

\def\eqref#1{equation~\ref{#1}}

\def\1{\bm{1}}

\DeclareMathAlphabet{\mathsfit}{\encodingdefault}{\sfdefault}{m}{sl}
\SetMathAlphabet{\mathsfit}{bold}{\encodingdefault}{\sfdefault}{bx}{n}

\usepackage{xspace}
\definecolor{searchblue}{HTML}{1A73E8}
\definecolor{searchlight}{HTML}{EAF1FB}
\definecolor{baselinegray}{HTML}{5F6368}
\definecolor{paperink}{HTML}{1A1A1A}
\definecolor{rulegray}{HTML}{D9D7D1}

\colorlet{querycolor}{searchblue}
\colorlet{responsecolor}{baselinegray}

\DeclareRobustCommand{\searchmark}{%
  \kern0.06em\raisebox{0.65ex}{%
    \includegraphics[height=1ex]{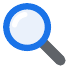}}}
\AtBeginDocument{\pdfstringdefDisableCommands{\def\searchmark{}}}

\DeclareRobustCommand{\method}{%
  \ifmmode\text{Search-Native Pretraining}%
  \else Search-Native Pretraining\fi\xspace}

\DeclareRobustCommand{\model}{%
  \ifmmode\text{QueryLM\searchmark}
  \else QueryLM\searchmark\fi\xspace}

\DeclareRobustCommand{\modelsize}[1]{%
  \ifmmode\text{QueryLM-#1\searchmark}%
  \else QueryLM-#1\searchmark\fi\xspace}

\newcommand{\baseline}{Standard\xspace}
\newcommand{\lmlm}{LMLM\xspace}
\newcommand{\colmlm}{Co-LMLM\xspace}

\DeclareRobustCommand{\hficon}{%
  \raisebox{-0.3ex}{\includegraphics[height=2.2ex]{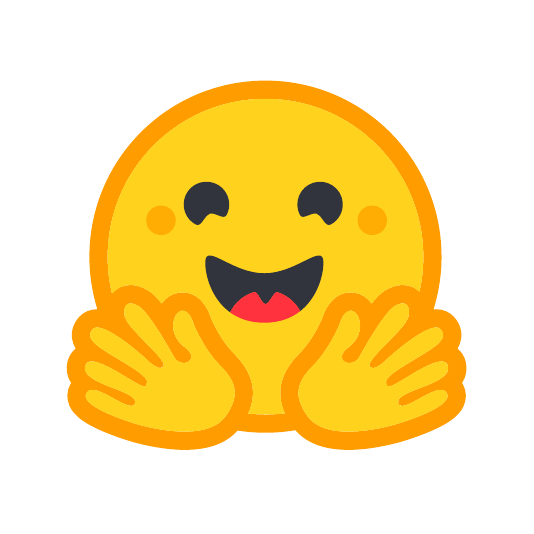}}}
\AtBeginDocument{\pdfstringdefDisableCommands{\def\hficon{Hugging Face}}}

\newcommand{\linkicon}[1]{\raisebox{-0.25ex}{\includegraphics[height=1.9ex]{figures/icons/#1.pdf}}}
\newcommand{\titlelinks}{{%
  \hypersetup{pdfborder={0 0 0}}
  \href{\projecturl}{\linkicon{search}\,\projectlabel}
  \qquad\href{\hfurl}{\hficon\,Models and Data}%
  \ifshowcodelink\qquad\href{\codeurl}{\linkicon{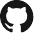}\,Code}\fi}}

\newcommand{\papertitlefont}{\normalfont\LARGE\centering}
\DeclareRobustCommand{\affiliationmark}[1]{%
  \raisebox{0.55ex}{\includegraphics[height=0.95ex]{figures/affiliations/#1.pdf}}}
\DeclareRobustCommand{\affilepfl}{\affiliationmark{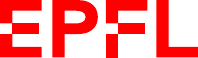}}
\DeclareRobustCommand{\affiltoronto}{%
  \raisebox{0.2ex}{\includegraphics[height=1.5ex]{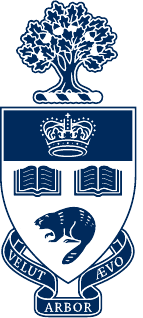}}}

\newcommand{\searchsft}{Search SFT\xspace}

\newcommand{\nosearch}{No search\xspace}

\newcommand{\wikipedia}{Wikipedia\xspace}
\newcommand{\dclmedu}{DCLM-Edu\xspace}
\newcommand{\sphere}{Sphere\xspace}
\newcommand{\smoltalk}{SmolTalk\xspace}
\newcommand{\smoltalktwo}{SmolTalk2\xspace}

\newcommand{\triviaqa}{TriviaQA\xspace}
\newcommand{\nqopen}{NQ-Open\xspace}
\newcommand{\popqa}{PopQA\xspace}
\newcommand{\webquestions}{WebQuestions\xspace}
\newcommand{\hotpotqa}{HotpotQA\xspace}
\newcommand{\twowikimultihopqa}{2WikiMultiHopQA\xspace}
\newcommand{\musique}{MuSiQue\xspace}
\newcommand{\bamboogle}{Bamboogle\xspace}
\newcommand{\freshqa}{FreshQA\xspace}

\newcommand{\piqa}{PIQA\xspace}

\newcommand{\halueval}{HaluEval\xspace}
\newcommand{\ragtruth}{RAGTruth\xspace}

\newcommand{\annotator}{\ensuremath{M_{\phi}}} 
\newcommand{\documentseq}{\ensuremath{x}}
\newcommand{\augdocument}{\ensuremath{\tilde{x}}}
\newcommand{\docprefix}{\ensuremath{p}}
\newcommand{\continuation}{\ensuremath{s}}
\newcommand{\query}{\ensuremath{q}}
\newcommand{\response}{\ensuremath{r}}

\newcommand{\insertions}{\ensuremath{\mathcal{A}}} 
\newcommand{\responsepositions}{\ensuremath{\mathcal{I_{R}}}}
\newcommand{\seeddata}{\ensuremath{\mathcal{D}_{\mathrm{seed}}}}
\newcommand{\insertionscore}{\ensuremath{\pi}}
\newcommand{\insertthreshold}{\ensuremath{\tau}}

\DeclareRobustCommand{\tooltoken}[1]{%
  \texttt{\detokenize{#1}}}

\newcommand{\modelhighlight}{%
  \textcolor{searchblue}{\textbf{\model}}}

\newcommand{\baselinehighlight}{%
  \textcolor{baselinegray}{\baseline}}

\newcommand{\searchrow}{\rowcolor{searchlight}}

\newcommand{\scoredelta}[1]{%
  \textsuperscript{\textcolor{baselinegray}{#1}}}

\newcommand{\ci}[1]{%
  \textsubscript{\textcolor{baselinegray!70}{$\pm$#1}}}

\newcommand{\lightrule}{\arrayrulecolor{rulegray}\midrule\arrayrulecolor{black}}
\newcommand{\grouprow}[2]{\rowcolor{rulegray!30}\multicolumn{#1}{l}{\textit{#2}} \\}

\usepackage{hyperref}
\usepackage{url}

\usepackage{subcaption}
\usepackage{graphicx}
\usepackage{booktabs}
\usepackage{adjustbox}
\usepackage{multirow}
\usepackage{placeins}
\usepackage{cleveref}
\usepackage{changepage}

\usepackage{listings}
\usepackage{tcolorbox}

\lstdefinestyle{prompt}{
    basicstyle=\ttfamily\footnotesize,
    columns=fullflexible,
    keepspaces=true,
    breaklines=true,
    breakatwhitespace=false,
    showstringspaces=false,
    frame=single
}

\lstdefinestyle{prompttext}{
    basicstyle=\ttfamily\footnotesize,
    columns=fullflexible,
    keepspaces=true,
    breaklines=true,
    showstringspaces=false,
    frame=none,
    aboveskip=0pt,
    belowskip=0pt
  }

\newtcolorbox{promptbox}[1]{
    title={#1},
    fonttitle=\footnotesize\bfseries,
    coltitle=black,
    colbacktitle=black!6,
    colback=black!2,
    colframe=black!30,
    boxrule=0.4pt,
    titlerule=0pt,
    arc=1mm,
    boxsep=0pt,
    left=6pt,right=6pt,
    top=4pt,bottom=4pt,
    toptitle=3pt,bottomtitle=3pt,
    before skip=0pt,
    after skip=5pt
  }
  
\lstdefinestyle{prompttextsafe}{
    style=prompttext,
    breaklines=true,
    breakatwhitespace=true,
    columns=fullflexible,
    keepspaces=true,
    showstringspaces=false,
    breakautoindent=false,
    breakindent=0pt
  }

\title{\papertitlefont Search From the Start\searchmark: Search-Native Pretraining for Language Models}

\author{%
\begin{minipage}[t]{\dimexpr\textwidth-2\tabcolsep\relax}
\centering\normalfont\normalsize
\textbf{Kaustubh Ponkshe\affilepfl\textsuperscript{*}\quad\thanks{Corresponding authors: 
    \texttt{kaustubh.ponkshe@epfl.ch} and \texttt{jayje@cs.toronto.edu}.}
Gyung Hyun Je\affiltoronto\textsuperscript{*}\quad
Stefan Taga\affilepfl\\[2pt]
Colin Raffel\affiltoronto\qquad Martin Jaggi\affilepfl}\\[5pt]
{\normalfont\small
\affilepfl\,EPFL\qquad
\affiltoronto\,University of Toronto\qquad
}\\[2pt]
{\normalfont\footnotesize\textsuperscript{*}Equal contribution.}
\end{minipage}%
}

\newif\ifpaperpreprint
\paperpreprinttrue
\ifpaperpreprint\iclrfinalcopy\fi
\begin{document}



\maketitle
\ifpaperpreprint\lhead{Preprint}\fi

\begin{abstract}
Modern language models (LMs) are increasingly deployed with access to search tools at inference. Yet pretraining remains predominantly focused on predicting static text, with search capabilities typically introduced during post-training.
We propose \method{} as a practical route to teach search capability during pretraining from ordinary text corpora.
Specifically, we augment pretraining corpus at scale with web-search queries and retrieve web response using a local index.
Training on this augmented corpus teaches models to generate search calls and incorporate retrieved evidence during language modeling.
Across \wikipedia{} and \dclmedu{} corpora and 360M to 3.3B parameter scales, search-pretrained models (\textbf{\model{}}) show improved performance on knowledge-intensive tasks and higher steerability toward the retrieved evidence.
Crucially, their advantage persists after supervised fine-tuning of both search-pretrained and conventionally pretrained models, indicating that the tested post-training procedures do not fully recover the capabilities developed during pretraining.
Throughout our experiments, we find that a \textbf{1.4B \model{} consistently beats or matches the performance of a FLOP-matched 3.3B baseline model} across all knowledge intensive tasks.
These results establish search-pretraining as a practical way to develop foundational search capabilities during pretraining.

\end{abstract}
\vspace{-0.6em}
{\centering\normalsize\titlelinks\par}  

\vfill  
\noindent\begin{minipage}{\linewidth}
  \centering
  \includegraphics[width=0.996\linewidth]{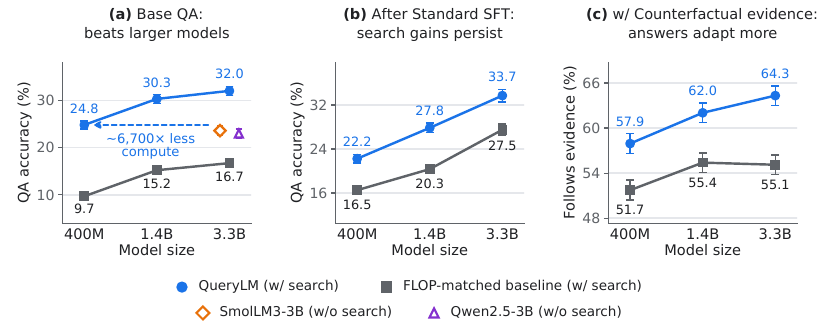}
  \vspace{-0.8em}
  \captionof{figure}{\textbf{\model{} is better at searching for and using evidence.}
  (a) After \method{}, \model{} with search outperforms the compute-matched baseline and the largest open-weight base models (SmolLM3-3B, Qwen2.5-3B) without search, despite their orders of magnitude more pretraining compute.
  (b) \model{} keeps higher accuracy on eight knowledge-intensive QA tasks after Standard SFT of both models.
  (c) After \searchsft{}, \model{} more often follows counterfactual evidence that contradicts established facts (LLM-judged answers on the same eight tasks).
  }
  \label{fig:teaser-scaling}
\end{minipage}
\vfill
\clearpage

\section{Introduction}

Language models (LMs) have finite capacity to store factual knowledge in their parameters \citep{morris2025languagemodelsmemorize,huang2026largermodelslearnmore}, which can limit their performance on knowledge-intensive tasks.
This poses a particular challenge for smaller models, as facts that cannot be inferred from general patterns must be memorized during pretraining \citep{allenzhu2024physicslanguagemodels33}.
Even when successfully encoded, knowledge in weights can become outdated as real-world facts change \citep{vu2023freshllms}.
Teaching an LM to search address these limitations by allowing factual knowledge to remain externally retrievable, with sources that can be updated without retraining the model.
Prior work has explored several ways to incorporate retrieved knowledge at inference time \citep{lazaridou2022internetaugmentedlanguagemodelsfewshot,nakano2022webgptbrowserassistedquestionansweringhuman,parisi2022talmtoolaugmentedlanguage,shuster2022blenderbot3deployedconversational,asai2023selfraglearningretrievegenerate,schick2023toolformerlanguagemodelsteach,yao2023reactsynergizingreasoningacting,jin2025searchr1trainingllmsreason}.
These capabilities are typically introduced during post-training \citep{openai2025gptoss120bgptoss20bmodel,dong2026agent,kang2026trace}, after the model has been pretrained to predict the next token without access to search.

However, recent work demonstrates the benefits of early exposure to capability-relevant data during pretraining \cite{zhang2025interplay,baek2026finetuner,minder2026synthetic}. Yet scaling such data to develop new capability during pretraining remains challenging, often requiring practitioners to rely on limited or synthetically generated data \cite{gunasekar2023textbooksneed,shao2024deepseekmath}.
We therefore ask: 

\begin{adjustwidth}{0.5cm}{0.5cm}
  \noindent\textit{``Can ordinary pretraining documents be augmented at scale
  to teach a downstream capability such as search, while preserving the original
  next-token prediction objective?''}
\end{adjustwidth}

In particular, we want an LM to use search to access diverse and evolving knowledge sources, such as the internet.
To teach the search ability during pretraining, we augment the training corpus with inline web-search tool calls and corresponding web-retrieved evidence that help predict the original document continuation.
These augmentations demonstrate when to seek information beyond the available text, what to search for, and how to use the returned evidence to predict the continuation. Together, they form our \method{} approach.

\begin{figure*}[t]
    \centering
    \includegraphics[width=\textwidth]{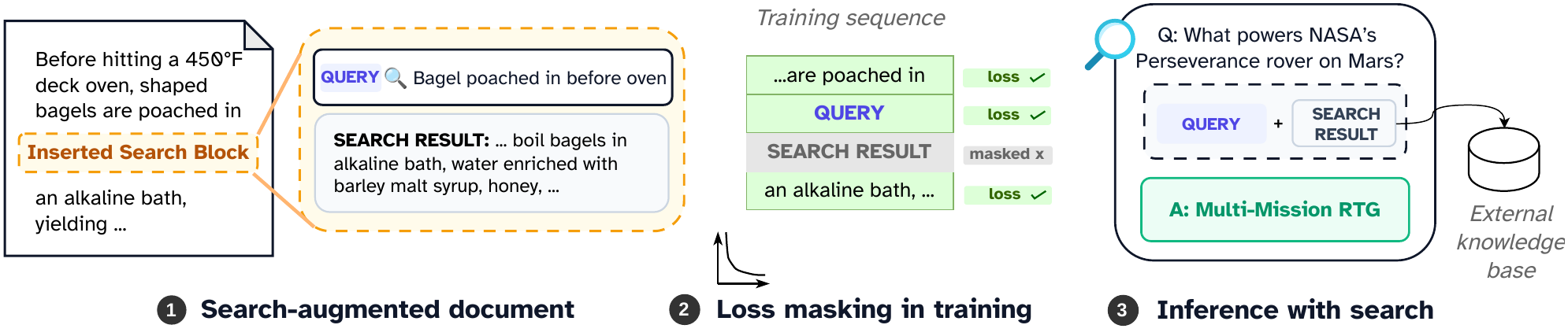}
    \caption{
    \textbf{\method{} pipeline overview.}
    \textbf{(1)} We augment training documents with inline search queries and corresponding search results, intended to contain useful knowledge for predicting original document continuation.
    \textbf{(2)} We train on \textbf{(1)}, computing loss only on original document tokens and search queries (green), while search results are masked (gray).
    \textbf{(3)} During inference, \model{} generates search queries to retrieve relevant information from an external knowledge base.
    }
    \label{fig:method_overview}
\end{figure*}



We construct search-augmented pretraining corpora from the knowledge-rich \wikipedia{} corpus and \dclmedu{}, which represents a realistic pretraining setting. The augmented \dclmedu{} corpus contains over 100B tokens of document text, search queries, and retrieved responses.
Using these corpora, we train \model{} models from scratch at scales from 360M to 3.3B parameters.
When search is available, \model{} improves on knowledge-intensive tasks compared with FLOP-matched baselines (\Cref{sec:results_base}).
Our 400M model with search also exceeds the average QA performance of SmolLM3-3B evaluated without search, despite using approximately 6,700 times less pretraining compute (\Cref{fig:teaser-scaling}).
The advantage over FLOP-matched baselines persists after both models undergo search-augmented fine-tuning, showing that the tested post-training recipe does not fully recover the benefit of learning search during pretraining (\Cref{sec:results_base}).
Beyond answering more questions correctly, \model{} more readily adjusts its predictions to supplied evidence, including evidence that contradicts facts seen during training (\Cref{sec:results_context}).
General capabilities remain broadly comparable after search fine-tuning (\Cref{sec:results_general}), while our ablations examine how retrieval quality, response length, and insertion frequency affect performance and training cost (\Cref{sec:discussion}).
These results support \method{} as a scalable approach to teaching models to search for and use external information from the start of training.

\section{Related Work}


\paragraph{LM knowledge retention.} Prior work has explored the limits of factual knowledge acquisition in language model parameters  \citep{kandpal2023largelanguagemodelsstruggle, morris2025languagemodelsmemorize}, highlighting challenges due to task feature interference \citep{huang2026largermodelslearnmore}, reduced model plasticity \citep{kim2025knowledgeentropydecaylanguage,springer2025overtrainedlanguagemodelsharder}, and forgetting \citep{liao2024exploringforgettinglargelanguage, luo2025empiricalstudycatastrophicforgetting}.
These challenges persists when the acquired knowledge becomes outdated and requires post-hoc editing \citep{lazaridou2021mindgapassessingtemporal,gu-etal-2024-model}.
Proposed mitigation includes data selection to prioritize learnable facts \citep{mindermann2022prioritizedtrainingpointslearnable, ye2026cramfitmoretraining}, modified parameter updates for reduced interference and easier editing \citep{wang2023orthogonalsubspacelearninglanguage,ghosal2025memorizationsinksisolatingmemorization}, increasing model plasticity \citep{chen2024improvinglanguageplasticitypretraining}, and complementing parametric knowledge with externally retrieved information (discussed below). 
Building on these efforts, we address the limitation of parametric knowledge by pretraining an LM to fill in the knowledge gap through search and ground its generation on retrieved passages.


\paragraph{External knowledge base.}
Prior work improves LMs on knowledge-intensive tasks by equipping them with external knowledge or tools. Inference-time or post-training approaches provide access to API-based tools or search engine during generation \citep{lazaridou2022internetaugmentedlanguagemodelsfewshot,nakano2022webgptbrowserassistedquestionansweringhuman,parisi2022talmtoolaugmentedlanguage,shuster2022blenderbot3deployedconversational,asai2023selfraglearningretrievegenerate,schick2023toolformerlanguagemodelsteach,yao2023reactsynergizingreasoningacting,jin2025searchr1trainingllmsreason}.
Among these work, Toolformer \citep{schick2023toolformerlanguagemodelsteach} is particularly relevant to our approach, as it injects tool call demonstrations inline and fine-tunes an LM to issue relevant tool calls.
Pretraining-focused efforts have explored jointly trained retrievers \citep{guu2020realmretrievalaugmentedlanguagemodel} or a specialized layer to attend to retrieved information \citep{borgeaud2022improvinglanguagemodelsretrieving}. More recent approaches, such as \lmlm{} \citep{zhao2025pretraininglimitedmemorylanguage} and \colmlm{} \citep{feldman2026colmlmcontinuousquerylimitedmemory}, build an explicit entity-based key-value stores from training or downstream task data and trains an LM to query these stores.
Unlike past methods, \model{} learns to issue flexible search queries during pretraining from scratch, without architectural modification or an explicit entity-based key-value store.

\paragraph{Pretraining data augmentation.} 
Previous studies have shown the benefits of introducing downstream capability-relevant data earlier during pretraining \citep{akter2025frontloadingreasoningsynergypretraining, zhou2026deepresearchpretrainingpredictive}, encoding desired behavior such as instruction-following ability \citep{cheng2024instructionpretraininglanguagemodels}, reduced toxicity \citep{maini2025safetypretraininggenerationsafe}, higher steerability \citep{gao2025metadataconditioningaccelerateslanguage}, improved reasoning \citep{akter2025mindmathinformedsynthetic, ruan2025reasoninglearnlatentthoughts}, knowledge synthesis \citep{huang2026citepretrainretrievalfreeknowledge}.
We similarly find that early exposure to search demonstrations in \method{} yields benefits that persist through post-training.
\section{\method}
\label{sec:search_augmented_pretraining}

\method{} trains a language model to issue search queries and incorporate retrieved evidence into subsequent generation. We construct demonstrations of these interactions by augmenting the training documents with search queries and externally retrieved evidence, while preserving the original document text.

\begin{figure}[b]
    \centering
    \includegraphics[width=\linewidth]{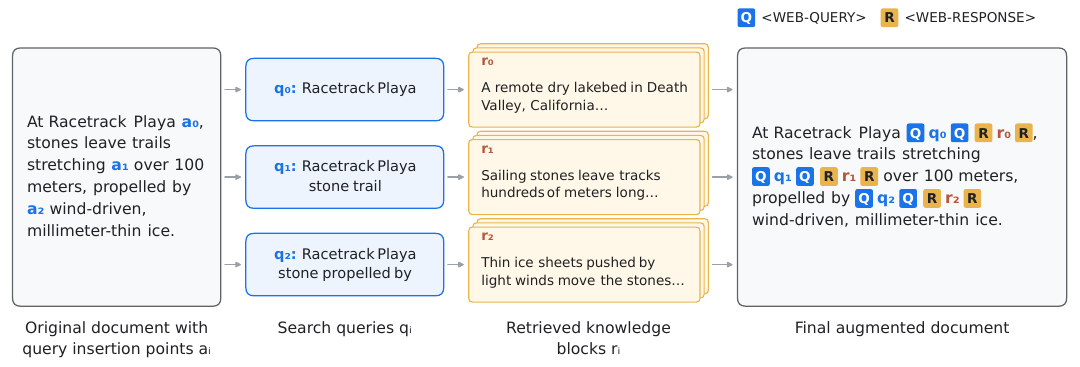}
    \caption{\textbf{Document annotation example} with inline search queries and retrieved documents.}
    \label{fig:annotated_text_example}
\end{figure}


\subsection{When and What to Search}
\label{sec:training_annotator_model}

For each document, we identify a set of search query insertion points $\insertions=\{a_1, ..., a_k \}$, each $a_i$ separating a prefix $\docprefix_i$, which sets the context motivating a search, from a suffix $\continuation_i$, the original document continuation.
A good insertion point follows the prefix $\docprefix_i$ that naturally motivates a search query $\query_i$, containing specific and relevant keywords from $\docprefix_i$ to retrieve useful context $\response_i$ (\Cref{fig:annotated_text_example}).
We denote the collections of queries and retrieved responses as $Q=\{\query_i\}_{i=1}^k$ and $R=\{\response_i\}_{i=1}^k$, respectively.


Extending prior work \citep{zhao2025pretraininglimitedmemorylanguage}, which places insertion points before an entity's associated value (e.g., between ``Dante Alighieri was born in'' and ``May 1265''), we consider more general insertion locations, such as immediately after a specific term mention without a linked value (e.g., placed after ``Dante Alighieri '') or positions where relevant information depends on distant context (e.g., between ``Dante Alighieri, [...], originally titled his masterwork'' and ``\textit{La Commedia}'').
We use an LM to identify suitable insertion points and generate corresponding search queries. To efficiently annotate the pretraining corpus at scale, we first fine-tune a small annotator model ($\annotator$) and then use the trained annotator to augment the full corpus.

\paragraph{Annotator model.}
We create a small search-augmented dataset ($\seeddata$) from an initial pool of 10,000 \wikipedia{} documents, annotated using GLM-5.2 \citep{glm5team2026glm5vibecodingagentic} as the teacher model.
We prompt the teacher to identify insertion points before specific facts that could benefit from external retrieval, such as names, dates, and quantities.

The teacher first identifies candidate insertion locations using the entire document but subsequently generates a search query given only the corresponding left context.
Query generation therefore uses the same available text as at inference.
To remove insertions with failed or empty search result, we retrieve and clean search responses for each search query (process detailed in \Cref{sec:search_response_retrieval}). 
Finally, we fine-tune Gemma 4 E4B \citep{gemmateam2026gemma4technicalreport} on $\seeddata$ to produce the search query annotator, $\annotator$. Further details on annotator training are given in \Cref{app:annotator_model_training_setup}.

\paragraph{Annotating pretraining corpus.}
For efficient search query insertion at pretraining scale, we process each sequence $\documentseq$ of length $T$ with $\annotator$, the fine-tuned annotatir model, in a single causal forward pass and compute query insertion scores following each $i$-th sequence token $\documentseq_i$ as
\[
\insertionscore_i(\documentseq) = \Pr\left(\tooltoken{<|WEB-QUERY>} \mid \documentseq_{1:i}\right), \quad i=1,\ldots,T,
\]
where \(\insertionscore_i(\documentseq)\) is the probability of inserting a search query immediately after token \(\documentseq_i\).
Positions with $\insertionscore_i(\documentseq)$ exceeding a threshold $\insertthreshold \in [0,1]$ form the final insertion set $\insertions=\{i\in\{1,\ldots,T\}\mid\insertionscore_i(\documentseq)\geq\insertthreshold\}=\{a_1,\ldots,a_k\}$.
Candidate insertion positions can therefore be scored in parallel within each document.
Our approach for sampling search insertion points refer to augmentation strategy in \cite{schick2023toolformerlanguagemodelsteach}.
Finally, $\annotator$ generates a query $\query_i$ for each $a_i$ given corresponding prefix $\docprefix_i$ until the closing delimiter. These queries are then passed to the search response retrieval pipeline (\Cref{sec:search_response_retrieval}).

\paragraph{Insertion filtering.}
For pretraining corpus, we use $\tau = 0.05$ as an initial filtering method to restrict the number of search queries inserted per document and drop low-quality search queries.
We then do additional filtering using ROUGE-1 F1 overlap between the retrieved response $\response_i$ and the original continuation $\continuation_i$, using lexical overlap as a proxy for relevance. 
The filtering threshold values and the filtering protocol are given in \Cref{app:annotating_pretraining_corpus}.

\subsection{Simulating Web Response During Pretraining}
\label{sec:search_response_retrieval}
To simulate a realistic and cost-efficient web-response, we use BM25 index over \sphere{} \citep{piktus2022weboysterknowledgeintensive}, a public dataset containing a broad collection of 134M web documents. We consider candidate indexes over different corpus based on retrieval coverage and storage requirements, detailed in \Cref{sec:retrieval_source_impact} and \Cref{app:context_retrieval}.

For each search query $\query_i$, we retrieve the top-10 scoring raw passages and select sentences with the most keyword stem overlap with the query, producing the web response $\response_i$ used in training corpus annotation.
The choice of how much retrieved information to retain introduces a trade-off: shorter search response reduces training overhead but may remove relevant information, and cleaning the raw passage using an LLM may help select more query-relevant information but incurs higher data preparation cost.
We evaluate the impact of training with web-response of different lengths and cleaning approach in \Cref{sec:training_context_influence} and \Cref{sec:context_length_query_frequency}.


\subsection{Pretraining Objective}
\label{sec:training_setup}

We train language models autoregressively on the annotated pretraining corpus. Given an annotated sequence $\augdocument$ of length $\widetilde{T}=|\augdocument|$ and indices $\responsepositions\subseteq\{1,\ldots,\widetilde{T}\}$ for web-response tokens (i.e., $\response_i$ and \tooltoken{<|WEB-RESPONSE|>} tokens), the training objective is
\[
\mathcal{L}(\theta)
=
-\sum_{t=1}^{\widetilde{T}}
\mathbf{1}[t\notin\responsepositions]
\log p_\theta(\augdocument_t\mid \augdocument_{<t}).
\]
The model therefore is trained with a next-token prediction objective on the original document text, search queries $\query_i$, and the special token \tooltoken{<|WEB-QUERY|>}, while masking the loss on web response tokens $\response_i$. This allows the model to condition the original document continuation on the web-response but avoids explicitly training it to reproduce the web response.

\section{Experimental Setup}
\label{sec:experiment}


\subsection{Pretraining Setup}
\label{sec:experiment_pretraining}

\paragraph{Pretraining corpora.}
We pretrain models using \method{} on two corpora: \wikipedia{} and \dclmedu{} \citep{allal2025smollm2smolgoesbig}.
\wikipedia{} provides a compact, knowledge-rich setting for controlled experiments. \dclmedu{} extends the study to more diverse web text, testing whether the benefits persist beyond encyclopedic documents. Both corpora use the annotation pipeline described in \Cref{sec:search_augmented_pretraining}, with the same
annotation model. See Appendix \Cref{app:pretraining_data} for data preprocessing details.

\paragraph{Baselines.}
For each of \model{} across varying scales, we pretrain a baseline model with the same architecture on original document text without search augmentation.
We approximately match each baseline's pretraining FLOPs to the corresponding \model{}, including computation on retrieved response tokens despite their exclusion from the loss during \method{} (see \Cref{app:pretraining_budgets} for detail).
We include \lmlm{} (360M) and \colmlm{} (400M) models off-the-shelf as our baselines. Our 360M-\model{} trained on \wikipedia{} provides the closest comparison with \lmlm{}, and our 400M-\model{} (overtrained) on \dclmedu{} is compared with \colmlm{} trained on FineWeb.

\paragraph{Model architectures and training budgets.}
For \textit{\wikipedia{}}, we train decoder-only Transformers from scratch with 360M parameters using the SmolLM2 configuration \citep{allal2025smollm2smolgoesbig}, to have the closest match to LMLM baseline. 
\wikipedia{} \model{} is trained on 33.4B tokens total over four epochs, with 4.01B tokens from original document tokens and 4.33B from search augmentation per epoch.

For \textit{\dclmedu{}}, we train nanochat configurations from scratch at 400M, 1.4B, and 3.3B parameters, excluding value embeddings. 
The 400M, 1.4B and 3.3B \model{}s are trained on 10.6B, 44.0B, and 99.8B tokens, 6.8B, 28.2B and 64.0B tokens from original document tokens (approx. 20 per parameter).
We include a deliberately over-trained 400M \dclmedu{} \model{} (65.1B document tokens, 101.5B processed) to support comparison with \colmlm{} trained on FineWeb. 
Further architecture details, optimization settings, and training budgets are given in \Cref{app:training_setup}.

\subsection{Supervised Fine-Tuning}
\label{sec:experiment_finetuning}

After pretraining, we fine-tune both \model{} and the FLOP-matched baselines to develop user-assistant conversation behavior. We use two different supervised fine-tuning (SFT) mixtures to answer different questions.
First, we perform standard SFT on a subsample of the Dolci dataset, which includes a search mixture, following the SFT setup of Olmo~3 \citep{olmo2026olmo3}.
Since many of the search traces in Dolci exceed our context length, we add our own search traces on \smoltalk{} conversations in an equal ratio.
This setup tests whether \model{} retains its advantage in the typical pipeline, where search is only introduced in post-training.
Second, we introduce \searchsft{}, which uses a much higher proportion of search-augmented data.
The \searchsft{} mixture contains 52,806 conversations from \smoltalk{} \citep{allal2025smollm2smolgoesbig} and \smoltalktwo{}, covering general chat skills and tool calling.
We additionally augment selected \smoltalk{} conversations with search interactions generated by our annotation model.
The main goal is to evaluate whether the advantages \method{} can be fully recovered in post-training with a search-heavy data distribution.
Further training details are provided in \Cref{app:finetuning_setup}. 

\subsection{Evaluation}
\label{sec:experiment_evaluation}

\paragraph{Tasks.}
We consider \triviaqa{} \citep{joshi2017triviaqa}, \nqopen{} \citep{kwiatkowski2019natural}, \popqa{} \citep{mallen2023popqa}, and \webquestions{} \citep{berant2013semantic} as factual knowledge probing tasks (``Factual QA''), as well as \hotpotqa{} \citep{yang2018hotpotqa}, \twowikimultihopqa{} \citep{ho2020constructingmultihopqadataset}, \musique{} \citep{trivedi-etal-2022-musique}, and \bamboogle{} \citep{press2023measuringnarrowingcompositionalitygap} for multi-hop question answering (``Multi-hop'') using the Search-R1 evaluation suite \citep{jin2025searchr1trainingllmsreason}. To assess factual grounding, we use on \freshqa{} \citep{vu2023freshllms} for time-sensitive factual question answering, and \halueval{} \citep{li2023haluevallargescalehallucinationevaluation} and \ragtruth{} \citep{niu2024ragtruthhallucinationcorpusdeveloping} for faithfulness to supplied evidence in hallucination-prone settings. Lastly, we evaluate \model{} on general language understanding and commonsense reasoning benchmarks. Task details are included in \Cref{app:evaluation_task}.

\paragraph{Prompting.}
The base models are given four-shot search demonstration for all QA tasks. \lmlm{} and \colmlm{} emit their own search queries to retrieve relevant values, and our FLOP-matched baseline uses its corresponding \model{}'s search queries and retrieved results. After \searchsft{}, both \model{} and FLOP-matched baselines are evaluated zero-shot and use their own search queries. Search is triggered for both base and SFT models with an injected \tooltoken{<|WEB-QUERY>} in the beginning of model generation.
Both base and SFT models are given budget of 1 look-up for factual QA and 2 look-ups for multi-hop QA.
We report \% correct answer is contained given the 24-token generation budget for base models, and LLM-judged accuracy with 64 token generation budget for SFT models. Prompt examples are included in \Cref{app:evaluation_prompting}. For each metric, we report the 95\% confidence intervals with bootstrap over questions.

\paragraph{Retrieval source.}
For fair comparison with \lmlm{}, which indexes its Wikipedia training corpus as the external knoweldge base, our \wikipedia{}-trained \model{} (360M) uses BM25 index over Wikipedia. Similarly, our \dclmedu{}-trained \model{} (400M) uses BM25 index over Fineweb-Edu and Wikipedia to match \colmlm{}'s retrieval corpus.
However, \model{} can flexibly switch to a web-like retrieval source, we report our main results including retrievals from Tavily, an API-based search engine. In \Cref{sec:retrieval_source_impact}, we demonstrate that switching to a more realistic retrieval source like Tavily improves the model performance for both \model{} and the FLOP-matched baselines.

\section{Results}
\label{sec:results}

\subsection{Search pretraining has advantage in knowledge intensive tasks}
\label{sec:results_base}


\begin{figure}[h]
    \centering
    \includegraphics[width=0.7\linewidth]{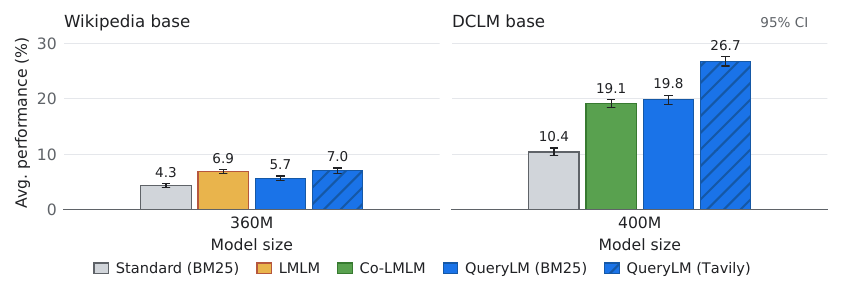}
    \caption{\textbf{Base model performance} on knowledge-intensive QA (avg. over eight tasks). 360M and 400M (overtrained) base \model{} with \lmlm{} and \colmlm{}. 
    }
    \label{fig:qa-base-baseline}
\end{figure}

\begin{figure}[h]
  \centering
  \includegraphics[width=\linewidth]{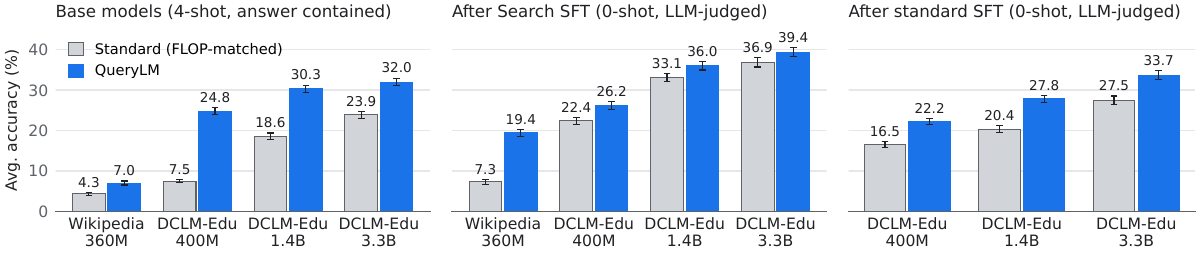}
  \caption{\textbf{Base and SFT model performance} on knowledge-intensive QA (avg. over eight tasks) across scale and training corpus.
   Left: base models.
   Middle: after \searchsft{}.
   Right: after standard SFT (Dolci mixture).
   Per-task scores in Appendix \Cref{tab:base-openqa-four-shot}, \Cref{tab:sft-qa}.
   }
  \label{fig:qa-base-sft}
\end{figure}

\paragraph{Base models.}
We evaluate \model{} on the eight knowledge-intensive QA tasks and report the average performance compared against our baselines.

\model{} outperforms both \lmlm{} and \colmlm{} when using Tavily as retrieval source (\Cref{fig:qa-base-baseline}).
With matched retrieval sources, \dclmedu{}-trained \model{} matches the performance of \colmlm{}. \wikipedia{}-trained \model{} slightly underperforms \lmlm{}, despite outperforming it on six of eight tasks, with exception being \popqa{} and WebQs. 
The structural similarity and potential factual overlap between \popqa{}, constructed from Wikidata knowledge triples, and \lmlm's Wikipedia-derived knowledge store may partially explain its strong performance on \popqa{}.
Across model scales and pretraining corpora, \model{} consistently outperforms the FLOP-matched baseline (\Cref{fig:qa-base-sft}, left).
Full results comparing open-weight base models and \model{} from \Cref{fig:teaser-scaling}-(a) are reported in Appendix, \Cref{tab:open-weight-closed-book}.

\paragraph{SFT models.}
Since our method can be replicated in post-training, we evaluate the advantage of search-pretraining by fine-tuning \model{} and FLOP-matched baselines on search-augmented data. With standard SFT, \model{} retains much of its base performance (\Cref{fig:qa-base-sft}, right).
Even under the \searchsft{} recipe, with a high proportion of search demonstrations, fine-tuned baselines do not fully recover the benefit of search pretraining, especially at smaller model scales (\Cref{fig:qa-base-sft}, middle).
We hypothesize that \method{} is particularly beneficial for smaller models, as learning to search and ground its predictions on retrieved results compensates for their limited parametric knowledge.
We further analyze the performance gap attributable to context grounding ability (\Cref{sec:results_context}) and search query writing quality (\Cref{sec:query_generation}).

\subsection{\model{} uses retrieved context to update its answers}
\label{sec:results_context}

To investigate whether \model{} grounds its predictions on retrieved evidence than on facts stored in its parameters, we measure its sensitivity to supplied evidence in a controlled setting, then evaluate whether the increased sensitivity translates to improvement on downstream tasks.

\paragraph{Updating knowledge based on counterfactual context.}

\begin{figure}[h]
    \centering

    \begin{subfigure}[t]{0.85\linewidth}
        \centering
        \includegraphics[width=\linewidth]{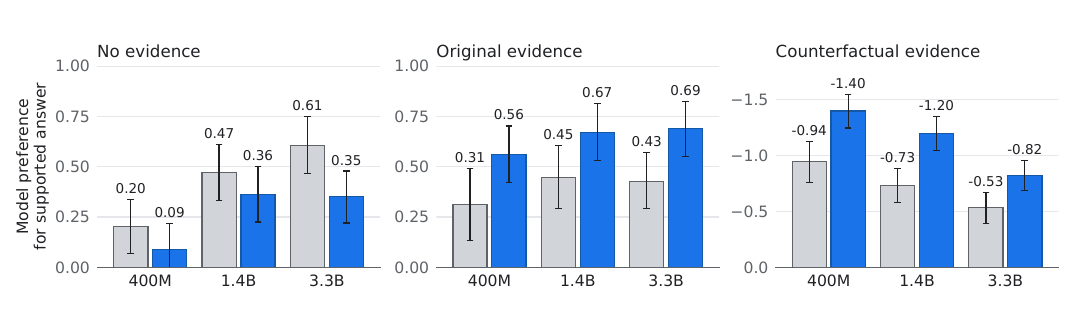}
        \caption{Model preference for facts seen during training. 
        Positive margins favor the original value, and negative margins favor the counterfactual.
        Counterfactual axis is reversed.}
        \label{fig:context-probe-training}
    \end{subfigure}

    \vspace{0.5em}

    \begin{subfigure}[t]{0.85\linewidth}
        \centering
        \includegraphics[width=\linewidth]{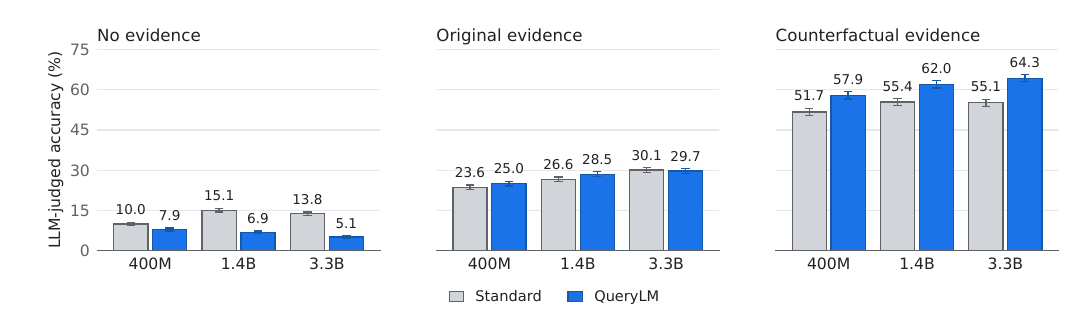}
        \caption{Model accuracy (LLM-judged) on the eight knowledge-intensive QA tasks.}
        \label{fig:context-probe-benchmark}
    \end{subfigure}

    \caption{\textbf{\dclmedu{}-trained SFT \model{}'s fact preference} under no evidence, supporting evidence, and counterfactual evidence settings. \textit{No evidence}: higher y-axis value indicates preference for original ground-truth facts. \textit{Original evidence} and \textit{Counterfactual evidence}: higher y-axis value indicates preference for facts supported by the evidence.}
    \label{fig:context-probe}
\end{figure}

We first sample 1000 stated facts in \dclmedu{} training documents and measure the model preference between ground-truth and plausible-but-wrong (``counterfactual'') values, under retrievals supporting either options.
For instance, we test if the counterfactual retrieval that falsely identifies Albert Einstein's birth-year as 1889 (correct: 1879) increases the model's preference for 1889.
We use GPT-oss-120B \citep{openai2025gptoss120bgptoss20bmodel} to edit the original ground-truth relevant evidence to instead support the counterfactual values. 
Model preference is measured using the log-probability difference between the original and counterfactual values, $\log P(\text{original}) - \log P(\text{counterfactual})$.

We find that both supporting and counterfactual evidence increase \model{}'s preference for the corresponding values more strongly than for the baseline models  (\Cref{fig:context-probe-training}). See \Cref{app:context_probe} for detailed setup and discussion.

\paragraph{Generation grounded on relevant evidence.}

We further evaluate \model{}'s context-following ability across the eight QA tasks using either the original ground-truth supporting and the edited counterfactual context (\Cref{fig:context-probe-benchmark}). To assess context following ability independently of query generation ability, we hold search queries fixed across \model{} and FLOP-matched baselines. Retrieval uses BM25 over Sphere. Details of setting are discussed in \Cref{app:context_probe}.

\model{} performs worse in closed-book setting, but is better at predicting the answer supported by supplied evidence, especially under counterfactual contexts.
These results suggest that \model{} relies more strongly on retrieved evidence. Additional evaluations on factual-groundedness benchmarks, including \freshqa{}, \halueval{}, and \ragtruth{}, show similar trends (Appendix, \Cref{app:complete_results}).




\subsection{General capabilities}
\label{sec:results_general}

For completeness, we evaluate \model{} on general language understanding and commonsense reasoning in closed-book setting (Appendix \Cref{app:general_capability}, \Cref{tab:general}).
\model{} base model slightly underperforms the FLOP-matched baselines, whereas the two are comparable after SFT, with mixed results across tasks.
Search pretraining preserves general capabilities, but its advantage on search-based tasks does not translate into consistent improvements in closed-book reasoning.


\section{Discussion}
\label{sec:discussion}

\subsection{How effective are \model{}'s search queries?}
\label{sec:query_generation}

We assess \model{} and FLOP-matched baselines' search query generation ability by comparing their downstream task performances using their own queries after \searchsft{} versus queries generated by a stronger model, Qwen3-30B-A3B-Instruct-2507 \citep{qwen3technicalreport}.
This allows us to assess the effectiveness of self-generated queries, and whether stronger query generation can narrow the performance gap between \model{} and baselines.
Both settings use BM25 over Sphere for retrieval.

\begin{figure}[ht]
    \centering
    \includegraphics[width=0.80\linewidth]{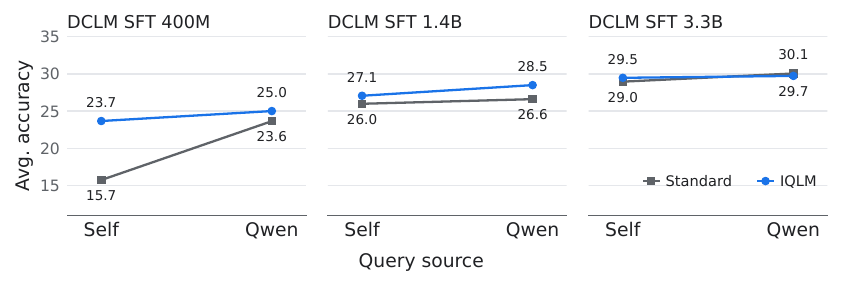}
    \caption{\textbf{Effect of query strength} on SFT models QA performance (avg. LLM-judged acc. (\%)).}
    \label{fig:sft-query-source}
\end{figure}

Across model scales, replacing self-generated queries with Qwen-generated queries results in smaller performance gains for \model{} than for the FLOP-matched baselines (\Cref{fig:sft-query-source}), suggesting that \model{} generates more effective search queries. 
When the query source is fixed, \model{} also outperforms the baselines at both 400M and 1.4B scales.
However, this trend reverses at the 3.3B scale, potentially due to stronger parametric knowledge and in-context reasoning ability at a larger scales.

\subsection{How does retrieval source affect \model{} performance?}
\label{sec:retrieval_source_impact}

\begin{figure}[t]
    \centering
    \includegraphics[width=0.9\linewidth]{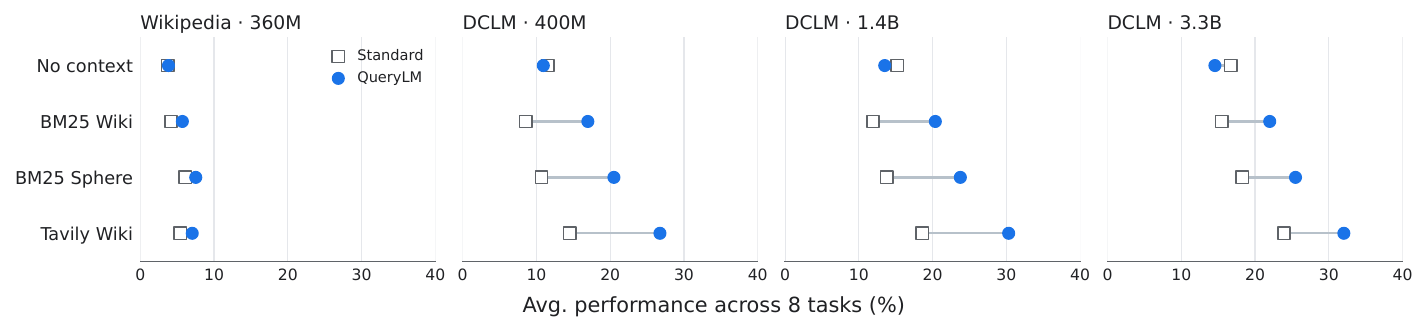}
    \caption{\textbf{Effect of retrieval source} on base model QA performance (avg. answer-contained score).}
    \label{fig:context-ablation}
\end{figure}

Using model-generated queries, we compare retrieval using BM25 over \wikipedia{}, BM25 over \sphere{}, and open-web search through Tavily. These settings vary in corpus coverage and retrieval backend, allowing us to assess how the choice of search source affects downstream accuracy.
Tavily gives the highest accuracy in this comparison, motivating its use for downstream evaluation. For pretraining data annotation, however, API costs accumulate over millions of documents. We therefore use local BM25 retrieval over \sphere{} during annotation. Raw values and trends for SFT models are in Appendix \Cref{app:retrieval_source_impact}.


\subsection{How does training-time search context affect performance?}
\label{sec:training_context_influence}

\begin{table*}[t]
  \centering
  \small
  \setlength{\tabcolsep}{3pt}
  \renewcommand{\arraystretch}{1.12}
  \begin{tabular}{l c c c c c c c}
  \toprule
  & & \multicolumn{3}{c}{Held-out document PPL} & \multicolumn{2}{c}{QA with search} & \\
  \cmidrule(lr){3-5}\cmidrule(lr){6-7}
  Pretraining context & Compute & No context & Orig. & Live & Open-domain & Multi-hop & General \\
  \midrule
  \baselinehighlight{} (none) & 1.00 & 7.57 & 10.01 & 9.70 & 3.2$^\dagger$ & 0.2$^\dagger$ & \textbf{35.1} \\
  \lightrule
  \searchrow \modelhighlight{} (default) & 1.04 & 7.96 & 7.44 & 7.78 & \textbf{12.2} & \textbf{5.7} & 33.7 \\
  \searchrow LM-summarized & 0.79 & 7.60 & 7.09 & 7.47 & 8.3 & 0.5 & 34.9 \\
  \searchrow Oracle & 1.04 & 10.17 & 5.03 & 10.07 & 9.7 & 4.0 & 31.8 \\
  \bottomrule
  \end{tabular}
  \caption{\textbf{Effect of the pretraining context} (\wikipedia{}, 360M; same documents and queries, only the response text differs). $^\dagger$baseline evaluated closed book.}
  \label{tab:ablation-training-context}
\end{table*}


We construct two search response variants for training data augmentation using an LM summarizer (Gemma 4 E4B): summaries of raw retrieved documents (``LM-summarized'') and summaries of the training documents themselves, which contain explicit ground-truth value asked for by the search query (``oracle'').
The former tests the benefit of cleaner retrieval context, while the latter examines the effect of training on highly relevant evidence and model's generalization beyond this controlled setting.
We compare both against the original \wikipedia{}-trained 360M \model{} and a FLOP-matched baseline (``Standard''). 

Without search context (''No context''), Oracle exhibits the highest perplexity on 400 held-out documents' original tokens, followed by \model{} and LM-summarized (\Cref{tab:ablation-training-context}). Perplexity drops slightly across all variants with the models' inference-time queries and context (``Live'').
With the original search annotations (``Orig.''), both Oracle and LM-summarized perplexity drops below \model{}'s, indicating greater reliance on the retrieved context. 
However, their stronger context reliance does not translate to improved performance on downstream QA (``QA with search'') and general benchmark (``General''), suggesting limited generalization to a different retrieval setting (BM25 retrieval over Sphere).



\subsection{Can we reduce training-time search response length and query frequency?}
\label{sec:context_length_query_frequency}

\begin{table*}[t!]
  \centering
  \small
  \setlength{\tabcolsep}{3pt}
  \renewcommand{\arraystretch}{1.12}
  \begin{tabular}{l c c c c c c c c}
  \toprule
  & & & \multicolumn{3}{c}{Held-out document PPL} & \multicolumn{2}{c}{QA with search} & \\
  \cmidrule(lr){4-6}\cmidrule(lr){7-8}
  Augmentation & Blocks/doc & Compute & No context & Gold & Live & Open-domain & Multi-hop & General \\
  \midrule
  \baselinehighlight{} (none) & -- & 1.00 & 7.57 & 10.01 & 9.70 & 3.2$^\dagger$ & 0.2$^\dagger$ & \textbf{35.1} \\
  \lightrule
  \searchrow \modelhighlight{} (default) & 8.2 & 1.04 & 7.96 & 7.44 & 7.78 & \textbf{12.2} & \textbf{5.7} & 33.7 \\
  \searchrow 50-char responses & 3.8 & 0.55 & 7.90 & 7.85 & 7.95 & 8.2 & 0.9 & 33.8 \\
  \searchrow Top-2 queries/doc & 2.0 & 0.57 & 7.94 & 7.73 & 7.90 & 5.6 & 1.2 & 33.5 \\
  \bottomrule
  \end{tabular}
  \caption{\textbf{Effect of context length and query frequency} (\wikipedia{}, 360M).
  Blocks/doc:
  inserted query--response blocks per held-out document. Compute: pretraining FLOPs relative to the compute-matched baseline.
  Other columns as in \Cref{tab:ablation-training-context}. $^\dagger$baseline evaluated closed-book.}
  \label{tab:ablation-length-frequency}
\end{table*}


Longer responses and more frequent searches increase the amount of retrieved information available during pretraining, but also increase training cost. We examine two ways to reduce this overhead: shortening the retrieved context and limiting the number of query insertions.

In the first variant, we limit the response length from 600 (our default) to 50 characters, which reduces the added sequence length and may also remove information useful for predicting the continuation. In the second variant, we retain only the two insertions per document with the highest response-continuation overlap. 
We find that both variants substantially reduce the amount of search augmentation while still retaining reasonable search capability useful for downstream performance (\Cref{tab:ablation-length-frequency}). These results suggest that context length and insertion frequency offer ways to trade performance for lower training cost.

As the Top-2 variant is the most practical and cost-effective alternative to our default augmentation, we test the approach at scale by training a FLOP-matched 3.3B $\model{}_{\text{Top-2}}$. On knowledge-intensive QA, $\model{}_{\text{Top-2}}$ performance falls between \model{} and \baseline{} after both \searchsft{} and standard SFT (\Cref{tab:sft-qa}), while on general capabilities it matches \baseline{} within the confidence intervals (\Cref{tab:general}). $\model{}_{\text{Top-2}}$ therefore offers a reasonable trade-off between downstream performance and annotation cost.



\subsection{What is the cost of \method{}?}
\label{sec:trade_off}

The annotation phase scans the training corpus of size $D$ and adds $A$-many queries of length $\ell_q$, adding $A\ell$ more tokens to training corpus, where $\ell$ is the size of query and response tokens. The data preparation is done once per corpus and can be reused across models.
During inference, cost scales linearly with the number of search-calls, adding $k \ell$ for $k$ rounds of search and retrievals. 
Number of search queries ($A$), tokens per query ($\ell$), and the number of queries at inference ($k$) are hyperparameters that can be tweaked to reduce cost, whose impact is discussed in \Cref{sec:training_context_influence} and \Cref{sec:context_length_query_frequency}. See \Cref{tab:cost_table} in Appendix \Cref{app:cost_method} for a summary table.

\section{Conclusion}

\method{} aims to teach LMs two complementary skills during pretraining: what to search for and how to use retrieved evidence to continue generation, alongside the standard language modelling objective. Our results show that \model{} acquires these capabilities, achieving higher performance on knowledge-intensive tasks with search and grounding its predictions more effectively in supplied evidence. The difference in the closed-book evaluations shows that the models are also successful in reframing much of the memorization in pretraining through the search objective. Together, these findings suggest that integrating search into pretraining can complement smaller models' parametric memory by training them to access and use external knowledge effectively.


\subsubsection*{Acknowledgments}
We acknowledge support by the Swiss AI Initiative Visiting PhD fellowships, as well as by the Cyber-Defence Campus, armasuisse S+T.

\bibliography{iclr2027_conference}
\bibliographystyle{iclr2027_conference}

\appendix
\section{Appendix}

\subsection{Data Annotation}

\paragraph{Creating the seed data.}
\label{app:model_annotation_strategy}

We construct $\seeddata$ from the first 10,000 English \wikipedia{} articles in the November 2023 snapshot that contain between 2,000 and 9,000 characters. 
GLM-5.2 annotates each document in two passes. It first reads the full document and marks where a search could help predict an upcoming fact, such as a name, date, or quantity. It then generates a search query at each position using only the preceding 1,600 characters (referred to as $p_i$ in the main text). This separates the choice of where to search from query generation, reducing the risk of including the answer in the query.

For each query, we retrieve passages with BM25 over FineWeb-Edu. We remove duplicate passages and passages for which at least 35\% of word trigrams also occur in the source article. We keep the five highest-ranked remaining passages, truncated to approximately 900 characters each. Gemma-4-E4B-it compresses them into one to three sentences. 
This seed-data pipeline uses LLM compression over FineWeb-Edu; the later pretraining corpus annotation uses simple CPU-based cleaning over \sphere{} for scalability.

We apply no additional relevance filter for queries: preliminary experiments showed that heavily filtering $\seeddata$ decreased the search query emission rate of the annotator. The final set contains approximately 20 search queries per document. We split train and validation data into 9:1 ratio for $\annotator$ fine-tuning.




\paragraph{Annotator model training setup.}
\label{app:annotator_model_training_setup}

We fine-tune Gemma 4 E4B-it \citep{gemmateam2026gemma4technicalreport} on the training split of $\seeddata$ using LoRA with rank 16, $\alpha=32$, and dropout 0.05. Adapters are applied to the text decoder's attention and feed-forward projections. We train for three epochs using AdamW, a peak learning rate of $2\times10^{-4}$, 3\% warmup, cosine decay, and an effective batch size of 32 sequences. Each training sequence is one annotated document, preceded by an instruction describing when and how to search. Sequences are truncated on the right at 4,096 tokens.

During tokenization, we map $\seeddata$'s query and response tags to Gemma's native tool tokens: \tooltoken{<|tool_call>} and \tooltoken{<tool_call|>} for queries, and \tooltoken{<|tool_response>} and \tooltoken{<tool_response|>} for responses.
Fine-tuning supervises the document text, queries, all four boundary tokens, and the end-of-turn marker, whereas the instruction prefix and retrieved response text are masked from the loss. Supervising the document text teaches the annotator to continue without searching where no call is present.
We use the final checkpoint after three epochs to annotate the pretraining corpora and search-SFT data. The same instruction prefix is used during annotation.


\paragraph{Annotating pretraining corpus.}
\label{app:annotating_pretraining_corpus}

We run $\annotator$ over the entire training corpus to identify query insertion points. We keep candidate positions with insertion probability $\tau \geq 0.05$, enforcing a minimum separation of 40 characters. Queries are generated using greedy decoding with a limit of 24 tokens. The same $\tau$ and query-generation budget are used for both \wikipedia{} and \dclmedu{}. 
After retrieving and compressing query responses, we measure their overlap with the first 260 characters of the original document continuation (referred to as $s_i$ in the main text) using ROUGE-1 F1. We retain insertions with a score of at least 0.1.



\paragraph{Choice of retrieval index.}
\label{app:context_retrieval}
We first evaluate the 100B and 1T-token subsets of FineWeb-Edu as retrieval corpora. Although scaling to 1T tokens improves the query response rate, the 100B-token subset already returns results for 90\% of search queries covered by 1T-token subset. We further find that \sphere{} \citep{piktus2022weboysterknowledgeintensive} retrieves more relevant results than the 100B-token FineWeb-Edu subset, as measured by the increase in log-likelihood of the original document continuation immediately following the retrieved results.

\subsection{Training Setup}
\label{app:training_setup}

\paragraph{Datasets and preprocessing.}
\label{app:pretraining_data}
For \wikipedia{}, we annotate the 6,083,581 English articles in the November 2023 snapshot (\texttt{20231101.en}). We restrict the corpus to the articles also present in \lmlm{}'s Wikipedia set and remove those without accepted search insertion. The final training corpus contains 4.01B original-document tokens and 8.43B tokens after search annotation. We reserve 1\% of training corpus as a held-out set.

For \dclmedu{}, we retain a subset of complete files from \texttt{HuggingFaceTB/dclm-edu}, retaining documents with educational scores of 2--5 without further score-based filtering. We remove documents shorter than 200 characters and exact duplicates within each file. The resulting corpus contains 65.13B original-document tokens and 101.54B tokens after search annotations. We reserve 1\% of training corpus as a held-out set.


Gathering retrieved responses for search annotation, we cap response length to approximately 400 characters for \wikipedia{} and 800 for \dclmedu{}. 
We pack documents into 4,096-token blocks using best-fit-decreasing packing, with a supervised end-of-text token after each document. 
Padding and retrieved responses are excluded from the loss; document text and queries are supervised.


\paragraph{Model architectures.}
\label{app:pretraining_architectures}
The \wikipedia{} models use the SmolLM2-360M configuration, with grouped-query attention, a SiLU feed-forward network, and tied input and output embeddings. The \dclmedu{} models use nanochat configurations with full multi-head attention, query/key normalization, squared-ReLU feed-forward networks, and untied input and output embeddings. They add gated value embeddings on alternating layers and cap output logits at 15. Wikipedia uses RMSNorm with $\epsilon=10^{-5}$; DCLM uses parameter-free RMSNorm. All configurations use rotary position embeddings with base 100,000 and omit biases. \Cref{tab:pretraining_architectures} gives the dimensions and parameter counts. The DCLM model sizes used throughout the paper exclude value embeddings; the table also reports total stored parameters.

All models use the SmolLM2 tokenizer extended with four search tokens, giving a vocabulary of 49,156 tokens. Nanochat pads the vocabulary dimension to 49,216. For the Wikipedia models, the four new embedding rows are initialized to the mean of the other rows.

\begin{table}[tbp]
\centering
\small
\setlength{\tabcolsep}{4pt}
\renewcommand{\arraystretch}{1.12}
\begin{tabular}{lrrrr}
\toprule
Corpus & \wikipedia{} & \multicolumn{3}{c}{\dclmedu{}} \\
Model size & 360M & 400M & 1.4B & 3.3B \\
\midrule
Layers & 32 & 14 & 24 & 32 \\
Hidden width & 960 & 1,280 & 2,048 & 2,816 \\
Feed-forward width & 2,560 & 5,120 & 8,192 & 11,264 \\
Query / key-value heads & 15 / 5 & 10 / 10 & 16 / 16 & 22 / 22 \\
Head dimension & 64 & 128 & 128 & 128 \\
\midrule
Reported parameters (M) & 361.8 & 401.2 & 1,409.6 & 3,322.3 \\
Value embeddings (M) & -- & 441.0 & 1,209.5 & 2,217.5 \\
Total stored parameters (M) & 361.8 & 842.2 & 2,619.1 & 5,539.7 \\
\bottomrule
\end{tabular}
\caption{\textbf{Pretraining model architectures}, shared by \model{} and \baseline{} within each comparison. 
}
\label{tab:pretraining_architectures}
\end{table}

\paragraph{Optimization.}
\label{app:pretraining_optimization}
\Cref{tab:pretraining_optimization} summarizes the common settings. We use a global batch size of 256 sequences, each containing 4,096 tokens.
The Wikipedia trainer processes all 4,096 positions per block; nanochat uses 4,095 input positions to predict the following tokens.
Wikipedia runs use fused AdamW with FlashAttention-2 without gradient checkpointing. DCLM runs use Muon for matrix weights and AdamW for input embeddings, value embeddings, and the output head, with bf16 computation and fp32 master weights.

For Muon, the effective peak learning rate is 0.028. Nesterov momentum increases from 0.85 to 0.97 over the first 400 steps, then decreases to 0.90 during learning-rate decay. We use five Polar Express iterations and weight decay 0.0096, annealed to zero with a cosine schedule during decay. 
For AdamW, the base learning rates are 0.3 for input embeddings and 0.008 for the output head, both scaled by $\sqrt{2}(d/768)^{-1/2}$ for hidden width $d$. Value embeddings use half the input-embedding learning rate. The AdamW coefficients $(\beta_1,\beta_2)$ are $(0.8,0.995)$ for input embeddings and $(0.8,0.96)$ for the output head. Weight decay is 0.001 for input embeddings and 0.01 for value embeddings and the output head.

The learning rate remains constant after warmup until the final decay phase.
Decay occupies the last 20\% of steps for both over-trained 400M models. The non-overtrained 400M \model{} and baseline pair branches from these runs at step 9,000, during the constant-learning-rate phase, and and decay over their final 1,100 steps (10.9\%). For the 1.4B models, decay spans 10.8\% and 6.8\% of training for the search and baseline models; and 3.5\% and 3.2\% for the 3.3B models respectively. These schedules therefore differ slightly within the larger compute-matched pairs.


\begin{table}[tbp]
\centering
\small
\setlength{\tabcolsep}{4pt}
\renewcommand{\arraystretch}{1.15}
\begin{tabular}{p{0.25\linewidth}p{0.29\linewidth}p{0.36\linewidth}}
\toprule
Setting & \wikipedia{} & \dclmedu{} \\
\midrule
Optimizer & AdamW & Muon + AdamW \\
Tokens per update & 1,048,576 & 1,048,320 \\
Peak learning rate & $5\times10^{-4}$ & 0.02828 (Muon); AdamW rates in text \\
Warmup & 2\% of steps & 40 steps, first segment only \\
Learning-rate decay & Cosine to zero & Linear to 5\% of peak \\
AdamW $\epsilon$ & $10^{-8}$ & $10^{-10}$ \\
Weight decay & 0.1 & Per optimizer/group; see text \\
AdamW $(\beta_1,\beta_2)$ & $(0.9,0.95)$ & Per parameter group; see text \\
Gradient-norm clipping & 1.0 & None \\
Precision & bf16 & bf16 compute, fp32 master weights \\
\bottomrule
\end{tabular}
\caption{\textbf{Pretraining optimization settings}. DCLM decay durations vary by run as specified in the text; all rates apply to pretraining, before supervised fine-tuning.}
\label{tab:pretraining_optimization}
\end{table}

\paragraph{Training budgets and compute matching.}
\label{app:pretraining_budgets}
\Cref{tab:pretraining_budgets} reports the final pretraining runs. ``Processed tokens'' include all input positions processed throughout training, including search queries and masked search responses. ``Document tokens'' count the original document tokens, also including repeated passes. 

The Wikipedia \model{} trains for four epochs and is compared with an eight-epoch baseline, using approximately 4\% more compute. 
The \dclmedu{}-trained \model{} at 400M, 1.4B, and 3.3B scales use 6.8B, 28.2B, and 64.0B document tokens, and their compute-matched baselines are trained on more document tokens to run for the same number of steps.
Matching compute requires the 3.3B baseline to repeat approximately 34.6B document tokens; the 400M and 1.4B baselines do not repeat documents. The over-trained 400M search model uses the full annotated corpus; its baseline runs 67 additional steps (0.07\%) and repeats approximately 36.5B document tokens.


\begin{table}[tbp]
    \centering
    \small
    \setlength{\tabcolsep}{3pt}
    \renewcommand{\arraystretch}{1.12}
    \begin{tabular}{l l l c c c c}
    \toprule
    & & & & \multicolumn{2}{c}{Tokens (B)} & FLOPs \\
    \cmidrule(lr){5-6}
    Corpus & Size & Condition & Steps & Document & Processed & ($10^{20}$) \\
    \midrule
    \searchrow \wikipedia{} & 360M & \modelhighlight{} & 31,860 & 16.0 & 33.4 & 0.631 \\
    & & \baselinehighlight{} & 30,600 & 32.1 & 32.1 & 0.606 \\
    \midrule
    \searchrow \dclmedu{} & 400M & \modelhighlight{} & 10,100 & 6.8 & 10.6 & 0.309 \\
    & & \baselinehighlight{} & 10,100 & 10.6 & 10.6 & 0.309 \\
    \lightrule
    \searchrow & 1.4B & \modelhighlight{} & 41,936 & 28.2 & 44.0 & 4.51 \\
    & & \baselinehighlight{} & 41,936 & 44.0 & 44.0 & 4.51 \\
    \lightrule
    \searchrow & 3.3B & \modelhighlight{} & 95,172 & 64.0 & 99.8 & 23.5 \\
    & & \baselinehighlight{} & 95,172 & 99.8 & 99.8 & 23.5 \\
    \midrule
    \searchrow \dclmedu{} & 400M, over-trained & \modelhighlight{} & 96,858 & 65.1 & 101.5 & 2.96 \\
    & & \baselinehighlight{} & 96,925 & 101.6 & 101.6 & 2.96 \\
    \bottomrule
    \end{tabular}
    \caption{\textbf{Pretraining budgets} before supervised fine-tuning.}
    \label{tab:pretraining_budgets}
\end{table}

\paragraph{Resources.}
Across \model{} and baseline model training, pretraining used approximately 316 GPU-hours on NVIDIA Hopper GPUs for \wikipedia{} (16$\times$H200) and 2,240 GPU-hours on NVIDIA Blackwell GPUs for \dclmedu{} (16$\times$B200).
These totals exclude annotation, ablations, and the four-epoch Wikipedia reference. GPU-hours sum training time across all devices.
  


\FloatBarrier
\subsection{Search Fine-Tuning}
\label{app:finetuning_setup}

\paragraph{Data mixture.}
\label{app:sft_data}
\model{} and \baseline{} are both fine-tuned on approximately 53,000 user-assistant conversations. 
Of these, about 64\% of data split consist of smol-rewrite, smol-summarize, smol-constraints, and everyday-conversations subsets of \smoltalk{} \citep{allal2025smollm2smolgoesbig}, teaching general chat skills and are not augmented with search annotation.
Another 15\% of data is Hermes function-calling and xLAM subsets of \smoltalktwo{} \citep{allal2025smollm2smolgoesbig}, demonstrating  tool calling behavior.
The remaining 21\% (11,237 conversations) consist of the Magpie, OpenHermes, and SystemChats subsets of \smoltalk{} and are augmented with search insertions.

We remove conversations for which web search is unlikely to be useful, such as fiction, mathematical calculations, code generation, or questions answerable from the prompt. We use the base Llama-3.1-8B model \citep{grattafiori2024llama3herdmodels}, prompted with labeled examples, as a binary classifier to make this decision.



\paragraph{Search annotation.}
\label{app:sft_search_annotation}
We annotate conversations with the same annotator, pipeline, and thresholds as for \dclmedu{} pretraining (\Cref{app:annotating_pretraining_corpus}),
restricting search augmentation to assistant turns. Search responses are retrieved with BM25 over \sphere{} and compressed using an 800-character budget.
After the initial annotation, we move all inserted search blocks to the start of the sentences in which they were originally placed, retaining at most one block per sentence. This enforces the behavior that an assistant searches before writing a claim rather than in the middle of it. Also, this setup matches our evaluation where search is triggered at the start of the assistant turn. 
The final data contain 38,436 search blocks across 9,713 conversations. The 1,524 conversations left without a block are kept as ordinary chat.


\paragraph{Format and loss masking.}
\label{app:sft_format}

Conversations are rendered as plain text, with \texttt{System:}, \texttt{User:}, and \texttt{Assistant:} role prefixes, with turns separated by blank lines. We add no additional chat tokens. 
Search blocks use the same query and response tokens as pretraining (\Cref{sec:training_setup}). 
We compute the loss on assistant text, search-query blocks, and end-of-sequence token, while masking and the loss on User turn, system prompts, retrieved responses, and padding. The loss on role prefixes are masked as well, except for the \texttt{User:} to signal end of assistant turn.


We pack conversations into 4,096-token sequences using best-fit-decreasing packing, truncating longer conversations. Attention is not masked between conversations in the same sequence. This gives 8,601 sequences with 35.2M tokens, 44\% of which carry loss. We reserve 1\% split consisting of 87 sequences as a held-out set.

\paragraph{Optimization.}
\label{app:sft_optimization}
Each corpus is fine-tuned for one epoch with its pretraining trainer (\Cref{tab:sft_optimization}). For \dclmedu{} models, we follow nanochat's fine-tuning recipe. Training resumes from the final pretraining checkpoint together with its optimizer state, including Muon momentum and AdamW moments; only the step counter and data position are reset. All learning rates are 0.8 times their pretraining peaks, and Muon momentum decreases from 0.97 to 0.90 as the learning rate decays. 
For \wikipedia{} models, we start a fresh AdamW optimizer.
Within each corpus, \model{} and \baseline{} use the same settings and number of updates. 



\begin{table}[tbp]
\centering
\small
\setlength{\tabcolsep}{4pt}
\renewcommand{\arraystretch}{1.15}
\begin{tabular}{p{0.25\linewidth}p{0.29\linewidth}p{0.36\linewidth}}
\toprule
Setting & \wikipedia{} & \dclmedu{} \\
\midrule
Initialization & Pretrained weights & Pretrained weights and optimizer state \\
Optimizer & AdamW & Muon + AdamW \\
Training length & 1 epoch, 134 updates & 1 epoch, 33 updates \\
Sequences per update & 64 & 256 \\
Tokens per update & 262,144 & 1,048,320 \\
Peak learning rate & $2\times10^{-5}$ & $0.8\times$ pretraining peaks (Muon 0.0226) \\
Warmup & 3\% of steps & None \\
Learning-rate decay & Cosine to zero & Constant for the first half, then linear to zero \\
Weight decay & 0.1 & 0 (Muon); pretraining values (AdamW) \\
\bottomrule
\end{tabular}
\caption{Search fine-tuning settings. Within each corpus, \model{} and \baseline{} use identical data and settings. Settings not listed follow \Cref{tab:pretraining_optimization}.}
\label{tab:sft_optimization}
\end{table}
\FloatBarrier





\subsection{Evaluation}
\label{app:evaluation}

\subsubsection{Tasks}
\label{app:evaluation_task}

We evaluate \model{} and baselines on eight open-ended question-answering tasks that probe world knowledge. Four tasks primarily assess factual recall: \textit{TriviaQA} covers broad factual knowledge through trivia questions; \textit{Natural Questions} (NQ-Open) pairs Google search questions with short answers; \textit{PopQA} tests knowledge derived from Wikipedia knowledge-triplets; and \textit{WebQuestions} contains question-answer pairs derived from web search queries.

The other four tasks assess multi-hop reasoning, requiring models to connect multiple pieces of evidence. \textit{HotpotQA} requires combining evidence across Wikipedia articles, while \textit{2Wiki} emphasizes explicit reasoning chains across Wikipedia facts.
\textit{MuSiQue} constructs questions requiring two to four reasoning steps by composing single-hop questions; we only consider two-hop questions from MuSiQue. \textit{Bamboogle} contains challenging two-hop questions whose answers are supported by Wikipedia articles.

In addition, we use \textit{\freshqa{}} to test whether retrieval helps modeles answer questions about current facts. Finally, \textit{\halueval{}} and \textit{\ragtruth{}} assess model's groundedness in supplied evidence in settings involving hallucinated or unsupported content.

\subsubsection{Evaluation budget}
\label{app:evaluation_budget}
In the search-enabled setting, a model makes one search query per question on the four factual QA tasks and up to two queries per question on the four multi-hop QA tasks. The first query is forced in both settings. Further search queries are suppressed for factual QA, while the second query is optional for multi-hop QA. We use greedy decoding for both base and SFT models, with a maximum of 24 tokens per search query. Answer generation is limited to 24 tokens for base models and 64 tokens for SFT models.

For each query, we retrieve the top-3 search results using either BM25 over the selected retrieval corpus or Tavily. We extract up to eight relevant sentences from the combined results, using a budget of approximately 1,200 characters that preserves complete sentences.

\subsubsection{Prompting}
\label{app:evaluation_prompting}
For base model evaluation, we use four-shot demonstrations of question and answer pairs sampled from the corresponding task. Closed-book evaluation contain regular question and answer pairs, whereas search-enabled demonstrations includes search and retrieval steps in each model's native format: explicit search query and result blocks for \model{}, fact blocks for \colmlm{}, and entity–relation look-ups for \lmlm{}. For multi-hop tasks, the four-shot demonstrations include both one-search and two-search examples.
SFT models are evaluated zero-shot, using ``User:/Assistant:'' format to be consistent with their training data. In the search-enable setting, the search query token is inserted at the start of the assistant response.
Example prompts are shown in \Cref{fig:prompt-formats}.

\begin{figure*}[h]

    \begin{promptbox}{\model{}: four-shot base model}
  \begin{lstlisting}[style=prompttextsafe]
  Question: What is the capital of Alaska?
  <WEB-QUERY>capital of Alaska</WEB-QUERY>
  <WEB-RESPONSE>[...] Juneau is the capital city in Alaska. [...]</WEB-RESPONSE>
  Answer: Juneau
  \end{lstlisting}
    \end{promptbox}

    \begin{promptbox}{\colmlm{}: four-shot base model}
  \begin{lstlisting}[style=prompttextsafe]
  Question: What is the capital of Alaska?
  Answer: <FACT>Juneau</FACT> Juneau
  \end{lstlisting}
    \end{promptbox}

    \begin{promptbox}{\lmlm{}: four-shot base model}
  \begin{lstlisting}[style=prompttextsafe]
  Question: What is the capital of Alaska?
  Answer: <|db_entity|> Alaska<|db_relationship|> Capital
  <|db_return|> Juneau<|db_end|> Juneau
  \end{lstlisting}
    \end{promptbox}

    \begin{promptbox}{\model{}: zero-shot SFT model}
  \begin{lstlisting}[style=prompttextsafe]
  User: Question: What is the capital of Alaska?
  Assistant: <WEB-QUERY>capital of Alaska</WEB-QUERY>
  <WEB-RESPONSE>[...] Juneau is the capital city in Alaska. [...]</WEB-RESPONSE>
  Juneau
  \end{lstlisting}
    \end{promptbox}
    \caption{Prompt formats for base and SFT models.}
    \label{fig:prompt-formats}
  \end{figure*}

\subsubsection{Impact of Multiple Look-ups on Knowledge-Intensive Task}
\label{app:evaluation_knowledge}

We find that allowing more than one search improves the model prediction accuracy for multi-hop questions. The benefits are greater for the SFT models (\Cref{tab:multihop-lookup-budget}).

\begin{table}[h!]
    \centering
    \small
    \setlength{\tabcolsep}{5pt}
    \begin{tabular}{l l c c c c c c}
    \toprule
    & & \multicolumn{3}{c}{Base (four-shot)}
      & \multicolumn{3}{c}{SFT (zero-shot)} \\
    \cmidrule(lr){3-5} \cmidrule(lr){6-8}
    Scale & Model
      & 1 lookup & $\leq 2$ lookups & $\Delta$
      & 1 lookup & $\leq 2$ lookups & $\Delta$ \\
    \midrule
    360M & \baseline
      & 3.64 & 3.64 & $0.00$
      & 6.59 & 6.54 & $-0.05$ \\
    \searchrow
         & \modelhighlight
      & 6.57 & 6.45 & $-0.13$
      & 10.84 & 11.24 & $+0.40$ \\
    \midrule
    400M & \baseline
      & 7.44 & 7.43 & $-0.01$
      & 12.41 & 12.51 & $+0.10$ \\
    \searchrow
         & \modelhighlight
      & 11.77 & 11.98 & $+0.21$
      & 15.31 & 16.82 & $+1.51$ \\
    \midrule
    1.4B & \baseline
      & 10.27 & 10.20 & $-0.06$
      & 18.31 & 18.98 & $+0.67$ \\
    \searchrow
         & \modelhighlight
      & 13.05 & 13.10 & $+0.05$
      & 18.56 & 20.22 & $+1.66$ \\
    \midrule
    3.3B & \baseline
      & 11.22 & 11.17 & $-0.05$
      & 21.50 & 21.68 & $+0.18$ \\
    \searchrow
         & \modelhighlight
      & 14.96 & 14.94 & $-0.02$
      & 20.21 & 21.69 & $+1.49$ \\
    \bottomrule
    \end{tabular}
    \caption{\textbf{Multi-hop QA for four-shot base and zero-shot SFT models} using BM25 retrieval over Sphere. Values are answer-containment accuracy (\%), averaged over Multi-hop QAs.}
    \label{tab:multihop-lookup-budget}
  \end{table}

\subsection{Complete Results}
\label{app:complete_results}

This section gives the full tables and setup behind the figures in \Cref{sec:results}, following its subsections.

\subsubsection{Knowledge-Intensive QA}
\label{app:qa_per_task}

In \Cref{tab:open-weight-closed-book}, we compare per-task performance and approximated pretraining FLOPs of open-weight models, SmolLM2 and 3 \citep{allal2025smollm2smolgoesbig, bakouch2025smollm3} and Qwen2.5 \citep{qwen2025qwen25technicalreport} base, with \model{} without and with search.
Search-enabled \model{} uses its own queries with Tavily open-web retrieval. The prompting strategy and generation budget follows the setup outlined in \Cref{app:evaluation}.

\begin{table}[h]
  \centering
  \small
  \setlength{\tabcolsep}{1.5pt}
  \renewcommand{\arraystretch}{1.12}
  \begin{adjustbox}{max width=\linewidth}
  \begin{tabular}{l r c c c c c c c c l}
  \toprule
  & & \multicolumn{4}{c}{Factual QA}
      & \multicolumn{4}{c}{Multi-hop QA} & \\
  \cmidrule(lr){3-6}\cmidrule(lr){7-10}
  Model & \shortstack{Pretraining\\FLOPs}
  & \triviaqa{} & NQ & \popqa{} & WebQs
  & \hotpotqa{} & 2Wiki & \musique{} & \bamboogle{} & Avg. \\
  \midrule
  \grouprow{11}{Open-weight base models: no search}
  SmolLM2-360M & $8.69{\times}10^{21}$
  & 26.9\ci{1.9} & 8.7\ci{1.2} & 11.2\ci{1.4} & 17.4\ci{1.7}
  & 12.4\ci{1.4} & 21.6\ci{1.8} & 2.7\ci{0.9} & 5.8\ci{4.3} & 13.3\ci{0.7} \\
  SmolLM2-1.7B & $1.13{\times}10^{23}$
  & 37.6\ci{2.1} & 17.2\ci{1.7} & 18.8\ci{1.7} & 25.4\ci{1.9}
  & 16.9\ci{1.6} & 22.4\ci{1.8} & 3.9\ci{1.1} & 9.9\ci{5.4} & 19.0\ci{0.9} \\
  SmolLM3-3B-Base & $2.07{\times}10^{23}$
  & 56.5\ci{2.2} & 23.2\ci{1.8} & 20.3\ci{1.8} & 27.2\ci{1.9}
  & 20.6\ci{1.8} & 25.3\ci{1.9} & 5.6\ci{1.3} & 9.9\ci{5.4} & 23.6\ci{0.9} \\
  \lightrule
  Qwen2.5-0.5B & $5.29{\times}10^{22}$
  & 20.4\ci{1.8} & 8.1\ci{1.2} & 11.8\ci{1.4} & 14.6\ci{1.5}
  & 14.2\ci{1.5} & 23.9\ci{1.9} & 2.5\ci{0.9} & 5.0\ci{4.1} & 12.5\ci{0.7} \\
  Qwen2.5-1.5B & $1.66{\times}10^{23}$
  & 38.4\ci{2.1} & 17.4\ci{1.7} & 14.8\ci{1.6} & 27.8\ci{2.0}
  & 18.1\ci{1.7} & 25.6\ci{1.9} & 3.9\ci{1.1} & 7.4\ci{4.8} & 19.2\ci{0.8} \\
  Qwen2.5-3B & $3.34{\times}10^{23}$
  & 44.3\ci{2.2} & 21.1\ci{1.8} & 17.0\ci{1.6} & 34.6\ci{2.1}
  & 21.6\ci{1.8} & 25.9\ci{1.9} & 6.0\ci{1.3} & 13.2\ci{6.0} & 23.0\ci{1.0} \\
  \midrule
  \grouprow{11}{\model{}: no search}
  \searchrow
  \modelhighlight{} 400M & $3.09{\times}10^{19}$
  & 12.2\ci{1.4} & 4.3\ci{0.9} & 8.2\ci{1.2} & 7.8\ci{1.2}
  & 9.2\ci{1.3} & 25.2\ci{1.9} & 0.6\ci{0.5} & 2.5\ci{3.1} & 8.8\ci{0.5} \\
  \searchrow
  \modelhighlight{} 1.4B & $4.51{\times}10^{20}$
  & 25.9\ci{1.9} & 9.5\ci{1.3} & 11.9\ci{1.4} & 18.9\ci{1.7}
  & 12.2\ci{1.4} & 20.9\ci{1.8} & 2.2\ci{0.8} & 6.6\ci{4.6} & 13.5\ci{0.7} \\
  \searchrow
  \modelhighlight{} 3.3B & $2.35{\times}10^{21}$
  & 31.3\ci{2.0} & 10.8\ci{1.4} & 12.2\ci{1.4} & 19.9\ci{1.7}
  & 12.6\ci{1.5} & 20.1\ci{1.8} & 3.0\ci{1.0} & 6.6\ci{4.6} & 14.6\ci{0.8} \\
  \midrule
  \grouprow{11}{\model{}: own-query open-web search via Tavily}
  \searchrow
  \modelhighlight{} 400M & $3.09{\times}10^{19}$
  & 59.2\ci{2.2} & 25.9\ci{1.9} & 37.3\ci{2.1} & 25.0\ci{1.9}
  & 19.9\ci{1.7} & 20.3\ci{1.8} & 3.8\ci{1.1} & 6.6\ci{4.6} & 24.8\ci{0.8} \\
  \searchrow
  \modelhighlight{} 1.4B & $4.51{\times}10^{20}$
  & 72.7\ci{2.0} & 33.7\ci{2.1} & 39.1\ci{2.1} & 34.2\ci{2.1}
  & 24.1\ci{1.9} & 22.2\ci{1.8} & 6.7\ci{1.4} & 9.9\ci{5.4} & 30.3\ci{0.9} \\
  \searchrow
  \modelhighlight{} 3.3B & $2.35{\times}10^{21}$
  & 74.5\ci{1.9} & 37.9\ci{2.1} & 42.8\ci{2.2} & 35.4\ci{2.1}
  & 25.4\ci{1.9} & 21.8\ci{1.8} & 7.9\ci{1.5} & 10.7\ci{5.6} & 32.0\ci{0.9} \\
  \bottomrule
  \end{tabular}
  \end{adjustbox}
  \caption{
      \textbf{Knowledge-intensive QA performance and approx. pretraining compute of open-weight base models}.
    (four-shot prompts, answer-contained accuracy (\%)).
    }
  \label{tab:open-weight-closed-book}
\end{table}
  
\Cref{tab:base-openqa-four-shot,tab:sft-qa} give the per-task scores behind \Cref{fig:qa-base-sft}.

\begin{table}[h!]
  \centering
  \small
  \setlength{\tabcolsep}{1.5pt}
  \renewcommand{\arraystretch}{1.12}
  \begin{adjustbox}{max width=\linewidth}
  \begin{tabular}{l l c c c c c c c c l}
  \toprule
  & & \multicolumn{4}{c}{Factual QA}
      & \multicolumn{4}{c}{Multi-hop QA} & \\
  \cmidrule(lr){3-6}\cmidrule(lr){7-10}
  Scale & Model
  & \triviaqa{} & NQ & \popqa{} & WebQs
  & \hotpotqa{} & 2Wiki & \musique{} & \bamboogle{} & Avg. \\
  \midrule
  \grouprow{11}{\wikipedia{} pretraining}
  360M & \baselinehighlight (BM25)
  & 6.4\ci{1.1} & 3.3\ci{0.8} & 3.3\ci{0.8} & 3.7\ci{0.8}
  & 1.5\ci{0.5} & 14.7\ci{1.6} & 1.0\ci{0.6} & 0.8\ci{2.2} & 4.3\ci{0.4} \\
  & \lmlm{} (382M)
  & 14.1\ci{1.5} & 5.1\ci{1.0} & \textbf{25.6}\ci{1.9} & \textbf{8.2}\ci{1.2}
  & 1.1\ci{0.5} & 0.6\ci{0.4} & 0.5\ci{0.4} & 0\ci{1.5} & 6.9\ci{0.4} \\
  \searchrow
  & \modelhighlight (BM25)
  & 10.7\ci{1.4} & 4.2\ci{0.9} & 5.1\ci{1.0} & 2.2\ci{0.6}
  & 5.4\ci{1.0} & \textbf{15.4}\ci{1.6} & 1.5\ci{0.7} & 0.8\ci{2.2}
  & 5.7\ci{0.4}\scoredelta{+1.4} \\
  \searchrow
  & \modelhighlight (Tavily)
  & \textbf{15.3}\ci{1.6} & \textbf{9.2}\ci{1.3} & 6.0\ci{1.0} & 3.1\ci{0.8}
  & \textbf{5.6}\ci{1.0} & 14.5\ci{1.5} & 0.9\ci{0.5} & \textbf{1.7}\ci{2.7}
  & \textbf{7.0}\ci{0.5}\scoredelta{+2.7} \\
  \midrule
  \grouprow{11}{\dclmedu{} 400M, over-trained (full annotated corpus, $\approx$10$\times$ compute)}
  400M & \baselinehighlight
  & 24.9\ci{1.9} & 8.7\ci{1.2} & 6.6\ci{1.1} & 9.7\ci{1.3}
  & 9.6\ci{1.3} & 16.8\ci{1.6} & 1.3\ci{0.6} & 5.8\ci{4.3} & 10.4\ci{0.7} \\
  & \colmlm{} (360M)
  & 51.7\ci{2.2} & 18.3\ci{1.7} & 30.4\ci{2.0} & 25.2\ci{1.9}
  & 8.4\ci{1.2} & 11.4\ci{1.4} & 3.5\ci{1.0} & 4.1\ci{3.8} & 19.1\ci{0.7} \\
  \searchrow
  & \modelhighlight (BM25)
  & 53.0\ci{2.2} & 17.7\ci{1.7} & 18.3\ci{1.7} & 19.7\ci{1.7}
  & 18.2\ci{1.7} & 21.4\ci{1.8} & 3.6\ci{1.0} & 6.6\ci{4.6}
  & 19.8\ci{0.8}\scoredelta{+9.4} \\
  \searchrow
  & \modelhighlight (Tavily)
  & \textbf{66.1}\ci{2.1} & \textbf{30.0}\ci{2.0} & \textbf{35.3}\ci{2.1} & \textbf{27.3}\ci{2.0}
  & \textbf{24.1}\ci{1.9} & \textbf{22.3}\ci{1.8} & \textbf{6.7}\ci{1.4} & \textbf{9.9}\ci{5.4}
  & \textbf{27.7}\ci{0.9}\scoredelta{+17.3} \\
  \midrule
  \grouprow{11}{\dclmedu{} pretraining (\model{} searches the open web using Tavily)}
  400M & \baselinehighlight
  & 21.9\ci{1.8} & 9.8\ci{1.3} & 5.5\ci{1.0} & 7.3\ci{1.1}
  & 2.6\ci{0.7} & 12.2\ci{1.4} & 0.7\ci{0.5} & 0.0\ci{1.5} & 7.5\ci{0.4} \\
  \searchrow
  & \modelhighlight
  & \textbf{59.2}\ci{2.2} & \textbf{25.9}\ci{1.9} & \textbf{37.3}\ci{2.1} & \textbf{25.0}\ci{1.9}
  & \textbf{19.9}\ci{1.7} & \textbf{20.3}\ci{1.8} & \textbf{3.8}\ci{1.1} & \textbf{6.6}\ci{4.6}
  & \textbf{24.8}\ci{0.8}\scoredelta{+17.3} \\
  \lightrule
  1.4B & \baselinehighlight
  & 47.9\ci{2.2} & 23.3\ci{1.9} & 23.9\ci{1.9} & 5.7\ci{1.0}
  & 16.1\ci{1.6} & 18.5\ci{1.7} & 6.0\ci{1.3} & 7.4\ci{4.8} & 18.6\ci{0.8} \\
  \searchrow
  & \modelhighlight
  & \textbf{72.7}\ci{2.0} & \textbf{33.7}\ci{2.1} & \textbf{39.1}\ci{2.1} & \textbf{34.2}\ci{2.1}
  & \textbf{24.1}\ci{1.9} & \textbf{22.2}\ci{1.8} & \textbf{6.7}\ci{1.4} & \textbf{9.9}\ci{5.4}
  & \textbf{30.3}\ci{0.9}\scoredelta{+11.7} \\
  \lightrule
  3.3B & \baselinehighlight
  & 58.1\ci{2.2} & 24.1\ci{1.9} & 33.3\ci{2.1} & 22.3\ci{1.8}
  & 19.8\ci{1.7} & 20.9\ci{1.8} & 5.7\ci{1.3} & 7.4\ci{4.8} & 23.9\ci{0.8} \\
  \searchrow
  & \modelhighlight
  & \textbf{74.5}\ci{1.9} & \textbf{37.9}\ci{2.1} & \textbf{42.8}\ci{2.2} & \textbf{35.4}\ci{2.1}
  & \textbf{25.4}\ci{1.9} & \textbf{21.8}\ci{1.8} & \textbf{7.9}\ci{1.5} & \textbf{10.7}\ci{5.6}
  & \textbf{32.0}\ci{0.9}\scoredelta{+8.1} \\
  \bottomrule
  \end{tabular}
  \end{adjustbox}
  \caption{\textbf{Base model knowledge-intensive QA evaluation} (four-shot, answer-contained scores (\%)). Deltas are average gains over the FLOP-matched baseline.
  }
 \label{tab:base-openqa-four-shot}
  \end{table}

\begin{table}[h!]
  \centering
  \small
  \setlength{\tabcolsep}{1.5pt}
  \renewcommand{\arraystretch}{1.12}
  \begin{adjustbox}{max width=\linewidth}
  \begin{tabular}{l l c c c c c c c c l}
  \toprule
  & & \multicolumn{4}{c}{Factual QA}
      & \multicolumn{4}{c}{Multi-hop QA} & \\
  \cmidrule(lr){3-6}\cmidrule(lr){7-10}
  Scale & Model
  & \triviaqa{} & NQ & \popqa{} & WebQs
  & \hotpotqa{} & 2Wiki & \musique{} & \bamboogle{} & Avg. \\
  \midrule
  \grouprow{11}{\wikipedia{} pretraining (\model{} searches \wikipedia{} using Tavily)}
  360M & \baselinehighlight
  & 12.6\ci{1.5} & 5.0\ci{1.0} & 11.6\ci{1.4} & 8.6\ci{1.2}
  & 6.8\ci{1.1} & 7.6\ci{1.2} & 1.8\ci{0.8} & 4.1\ci{3.8} & 7.3\ci{0.6} \\
  \searchrow
  & \modelhighlight
  & \textbf{37.1}\ci{2.1} & \textbf{18.6}\ci{1.7} & \textbf{28.8}\ci{2.0} & \textbf{23.5}\ci{1.9}
  & \textbf{19.6}\ci{1.7} & \textbf{15.9}\ci{1.6} & \textbf{4.9}\ci{1.2} & \textbf{6.6}\ci{4.6}
  & \textbf{19.4}\ci{0.8}\scoredelta{+12.1} \\
  \midrule
  \grouprow{11}{\dclmedu{} pretraining (\model{} searches the open web using Tavily)}
  400M & \baselinehighlight
  & 39.2\ci{2.1} & 22.1\ci{1.8} & 30.8\ci{2.0} & 32.7\ci{2.1}
  & 19.0\ci{1.7} & 20.0\ci{1.8} & 6.9\ci{1.4} & 8.3\ci{5.0} & 22.4\ci{0.9} \\
  \searchrow
  & \modelhighlight
  & \textbf{41.8}\ci{2.2} & \textbf{27.5}\ci{2.0} & \textbf{37.5}\ci{2.1} & \textbf{37.3}\ci{2.1}
  & \textbf{21.9}\ci{1.8} & \textbf{23.5}\ci{1.9} & \textbf{7.6}\ci{1.5} & \textbf{12.4}\ci{5.9}
  & \textbf{26.2}\ci{1.0}\scoredelta{+3.8} \\
  \lightrule
  1.4B & \baselinehighlight
  & 64.9\ci{2.1} & 37.0\ci{2.1} & 37.4\ci{2.1} & 48.9\ci{2.2}
  & 29.8\ci{2.0} & 25.1\ci{1.9} & 10.4\ci{1.7} & 11.6\ci{5.7} & 33.1\ci{1.0} \\
  \searchrow
  & \modelhighlight
  & \textbf{70.0}\ci{2.0} & \textbf{41.9}\ci{2.2} & \textbf{40.7}\ci{2.2} & \textbf{50.7}\ci{2.2}
  & \textbf{31.3}\ci{2.0} & \textbf{28.1}\ci{2.0} & \textbf{11.9}\ci{1.8} & \textbf{13.2}\ci{6.0}
  & \textbf{36.0}\ci{1.0}\scoredelta{+2.9} \\
  \lightrule
  3.3B & \baselinehighlight
  & 70.2\ci{2.0} & 40.5\ci{2.1} & 39.3\ci{2.1} & 52.0\ci{2.2}
  & 32.1\ci{2.0} & 24.4\ci{1.9} & 12.6\ci{1.8} & \textbf{24.0}\ci{7.5} & 36.9\ci{1.2} \\
  \searchrow
  & \modelhighlight
  & \textbf{74.4}\ci{1.9} & \textbf{46.3}\ci{2.2} & \textbf{40.8}\ci{2.2} & \textbf{54.0}\ci{2.2}
  & \textbf{34.3}\ci{2.1} & \textbf{29.3}\ci{2.0} & \textbf{14.1}\ci{1.9} & 22.3\ci{7.4}
  & \textbf{39.4}\ci{1.2}\scoredelta{+2.5} \\
  & \textcolor{searchblue}{$\model{}_{\text{Top-2}}$}
  & 66.8\ci{2.1} & 40.7\ci{2.2} & 42.6\ci{2.2} & 50.4\ci{2.2}
  & 32.2\ci{2.0} & 31.2\ci{2.0} & 12.7\ci{1.8} & 26.4\ci{7.8}
  & 37.9\ci{1.2}\scoredelta{+1.0} \\
  \midrule
  \grouprow{11}{\dclmedu{} pretraining, standard SFT (Dolci mixture; \model{} searches the open web using Tavily)}
  400M & \baselinehighlight
  & 30.2\ci{2.0} & 15.2\ci{1.6} & 21.6\ci{1.8} & 25.7\ci{1.9}
  & 13.4\ci{1.5} & 16.2\ci{1.6} & 4.0\ci{1.1} & \textbf{5.9}\ci{4.4} & 16.5\ci{0.8} \\
  \searchrow
  & \modelhighlight
  & \textbf{39.5}\ci{2.1} & \textbf{27.0}\ci{1.9} & \textbf{33.9}\ci{2.1} & \textbf{33.2}\ci{2.1}
  & \textbf{15.1}\ci{1.6} & \textbf{18.8}\ci{1.7} & \textbf{5.3}\ci{1.3} & 5.1\ci{4.2}
  & \textbf{22.2}\ci{0.8}\scoredelta{+5.7} \\
  \lightrule
  1.4B & \baselinehighlight
  & 42.9\ci{2.2} & 21.1\ci{1.8} & 20.4\ci{1.8} & 32.8\ci{2.1}
  & 15.8\ci{1.6} & 18.8\ci{1.7} & 4.4\ci{1.1} & 6.6\ci{4.6} & 20.4\ci{0.8} \\
  \searchrow
  & \modelhighlight
  & \textbf{55.0}\ci{2.2} & \textbf{32.0}\ci{2.0} & \textbf{35.1}\ci{2.1} & \textbf{42.1}\ci{2.2}
  & \textbf{23.2}\ci{1.8} & \textbf{20.3}\ci{1.8} & \textbf{6.7}\ci{1.4} & \textbf{8.3}\ci{5.0}
  & \textbf{27.8}\ci{0.9}\scoredelta{+7.4} \\
  \lightrule
  3.3B & \baselinehighlight
  & 52.4\ci{2.2} & 27.4\ci{2.0} & 30.3\ci{2.0} & 39.3\ci{2.1}
  & 23.2\ci{1.8} & \textbf{23.7}\ci{1.9} & 7.9\ci{1.5} & 15.7\ci{6.5} & 27.5\ci{1.0} \\
  \searchrow
  & \modelhighlight
  & \textbf{61.8}\ci{2.1} & \textbf{39.5}\ci{2.1} & \textbf{39.5}\ci{2.1} & \textbf{46.9}\ci{2.2}
  & \textbf{27.9}\ci{2.0} & 22.9\ci{1.8} & \textbf{8.7}\ci{1.6} & \textbf{22.3}\ci{7.4}
  & \textbf{33.7}\ci{1.1}\scoredelta{+6.2} \\
  & \textcolor{searchblue}{$\model{}_{\text{Top-2}}$}
  & 58.9\ci{2.2} & 37.3\ci{2.1} & 38.1\ci{2.1} & 49.5\ci{2.2}
  & 28.6\ci{2.0} & 26.1\ci{1.9} & 8.5\ci{1.6} & 13.2\ci{6.0}
  & 32.5\ci{1.0}\scoredelta{+5.0} \\
  \bottomrule
  \end{tabular}
  \end{adjustbox}
  \caption{\textbf{SFT models knowledge-intensive QA evaluation} (zero-shot, LLM-judged accuracy). Multi-hop QA allows up to two searches. Unless marked as standard SFT, models are trained with \searchsft{}. 
  $\model{}_{\text{Top-2}}$: compute-matched 3.3B \model{}, pretrained with only the two highest-overlap search insertions per document (\Cref{sec:context_length_query_frequency}); 
  Deltas are average gains over the FLOP-matched baseline. 
  }
  \label{tab:sft-qa}  
  \end{table}

\subsubsection{Use of Retrieved Context}
\label{app:context_probe}

\paragraph{Counterfactual Context.}

To compare \model{} and the baseline's ability to follow counterfactual evidence, we consider \searchsft{}-trained models as they are both trained with inline search demonstrations.

For model preference shift evaluation, we sample 1000 sequences from \dclmedu{} training documents that \model{} was trained on with search annotations, with search query and retrieval that support the original continuation. We filter for search insertions that are immediately followed by a concrete fact.
We then measure the log-probability difference of the ground-truth and counterfactual values at the same continuation position under under three conditions: no evidence, cleaned original evidence, and counterfactual evidence.

For counterfactual QA evaluation, we use Qwen3-30B-A3B-Instruct-2507 model generated search queries to use the same retrieval source for both \model{} and baseline models. We use BM25 index over Sphere for reproducible retrieval. The model performance is judged by GPT-oss-120B, following the same SFT performance evaluation protocol.

\Cref{tab:context-probe} gives the values behind \Cref{fig:context-probe}, for both the base and SFT models. Since \baseline{} models have not seen the search query annotations during training, we also assess the preference shift by placing the retrieved context at the start of the input sequence, instead of right before the factual value (``context first'').

\begin{table}[h!]
  \centering
  \small
  \setlength{\tabcolsep}{4.5pt}
  \renewcommand{\arraystretch}{1.12}
  \begin{adjustbox}{max width=\linewidth}
  \begin{tabular}{l r r r r r r r r r}
  \toprule
  & \multicolumn{3}{c}{No evidence} & \multicolumn{3}{c}{Original evidence} & \multicolumn{3}{c}{Counterfactual evidence} \\
  \cmidrule(lr){2-4}\cmidrule(lr){5-7}\cmidrule(lr){8-10}
  Model & 400M & 1.4B & 3.3B & 400M & 1.4B & 3.3B & 400M & 1.4B & 3.3B \\
  \midrule
  \grouprow{10}{Base models}
  \baselinehighlight & 0.20\ci{0.12} & 0.38\ci{0.19} & 0.49\ci{0.18} & 0.35\ci{0.14} & 0.43\ci{0.18} & 0.48\ci{0.17} & $-$1.20\ci{0.13} & $-$0.89\ci{0.17} & $-$1.00\ci{0.16} \\
  \baselinehighlight{} (context first) & 0.20\ci{0.12} & 0.38\ci{0.19} & 0.49\ci{0.19} & 0.67\ci{0.13} & 0.79\ci{0.20} & 0.89\ci{0.19} & $-$1.29\ci{0.13} & $-$0.99\ci{0.19} & $-$0.91\ci{0.20} \\
  \searchrow
  \modelhighlight & 0.15\ci{0.11} & 0.33\ci{0.19} & 0.40\ci{0.19} & 0.66\ci{0.13} & 0.81\ci{0.20} & 0.89\ci{0.20} & $-$1.27\ci{0.14} & $-$0.91\ci{0.20} & $-$0.75\ci{0.20} \\
  \midrule
  \grouprow{10}{After \searchsft{}}
  \baselinehighlight & 0.17\ci{0.14} & 0.25\ci{0.18} & 0.38\ci{0.19} & 0.25\ci{0.19} & 0.35\ci{0.17} & 0.46\ci{0.17} & $-$1.07\ci{0.19} & $-$0.82\ci{0.17} & $-$0.57\ci{0.17} \\
  \searchrow
  \modelhighlight & 0.12\ci{0.13} & 0.21\ci{0.17} & 0.30\ci{0.17} & 0.63\ci{0.15} & 0.82\ci{0.18} & 0.78\ci{0.18} & $-$1.46\ci{0.15} & $-$1.47\ci{0.20} & $-$0.93\ci{0.18} \\
  \bottomrule
  \end{tabular}
  \end{adjustbox}
  \caption{\textbf{Controlled context probe} on 1000 held-out \dclmedu{} facts. 
  }
  \label{tab:context-probe}
\end{table}

\paragraph{Model Groundedness.}

We evaluate \model{} and \baseline{} models on \freshqa{}, \halueval{}, and \ragtruth{} for groundedness evaluation.

We evaluate factual accuracy on \freshqa{} using the 500 test questions from the 21 April 2026 snapshot. The benchmark covers facts with varying rates of change (``Fact type'') and questions with false premises (``Premise''), which tests access to current knowledge and the ability to reject incorrect assumptions.
Before SFT, we compare search-enabled \model{} with the closed-book \baseline{} model; after \searchsft{}, both models are evaluated with search enabled.
\Cref{tab:freshqa} shows \model{} substantially improves over the closed-book baseline before SFT. After search SFT, the two models perform similarly at every scale.

\begin{table}[h!]
  \centering
  \small
  \setlength{\tabcolsep}{3.3pt}
  \renewcommand{\arraystretch}{1.12}
  \begin{adjustbox}{max width=\linewidth}
  \begin{tabular}{l l c c c c c c l}
  \toprule
  & & & \multicolumn{3}{c}{Fact type}
      & \multicolumn{2}{c}{Premise} & \\
  \cmidrule(lr){4-6}\cmidrule(lr){7-8}
  Scale & Model & \nosearch
    & Never-changing & Slow-changing & Fast-changing
    & False & Valid & Avg. \\
  \midrule
  \grouprow{9}{Base models}
  400M & \baselinehighlight
    & 5.8\ci{2.1} & 9.0\ci{4.1} & 2.2\ci{2.3} & 6.1\ci{4.2} & 0.0\ci{1.5} & 7.7\ci{2.7} & 5.8\ci{2.1} \\
  \searchrow
       & \modelhighlight
    & \textbf{6.0}\ci{2.1}
    & \textbf{34.6}\ci{6.7} & \textbf{17.8}\ci{5.6} & \textbf{9.1}\ci{5.0}
    & \textbf{6.5}\ci{4.5} & \textbf{26.9}\ci{4.5}
    & \textbf{21.8}\ci{3.6}\scoredelta{+16.0} \\
  \lightrule
  1.4B & \baselinehighlight
    & \textbf{10.2}\ci{2.7}
    & 17.6\ci{5.4} & 6.1\ci{3.6} & 5.3\ci{4.0} & 0.0\ci{1.5} & 13.6\ci{3.5} & 10.2\ci{2.7} \\
  \searchrow
       & \modelhighlight
    & 8.4\ci{2.4}
    & \textbf{42.6}\ci{7.0} & \textbf{22.2}\ci{6.0} & \textbf{9.1}\ci{5.0}
    & \textbf{4.8}\ci{3.9} & \textbf{33.5}\ci{4.7}
    & \textbf{26.4}\ci{3.9}\scoredelta{+16.2} \\
  \lightrule
  3.3B & \baselinehighlight
    & \textbf{13.2}\ci{3.0}
    & 23.9\ci{6.1} & 8.3\ci{4.1} & 4.5\ci{3.7} & \textbf{4.0}\ci{3.7} & 16.2\ci{3.7} & 13.2\ci{3.0} \\
  \searchrow
       & \modelhighlight
    & 8.8\ci{2.5}
    & \textbf{42.6}\ci{7.0} & \textbf{24.4}\ci{6.2} & \textbf{13.6}\ci{5.9}
    & 3.2\ci{3.4} & \textbf{36.7}\ci{4.8}
    & \textbf{28.4}\ci{3.9}\scoredelta{+15.2} \\
  \midrule
  \grouprow{9}{After \searchsft{}}
  400M & \baselinehighlight
    & 2.4\ci{1.4}
    & 28.7\ci{6.4} & 18.9\ci{5.7} & 6.1\ci{4.2} & \textbf{6.5}\ci{4.5} & 23.4\ci{4.3} & 19.2\ci{3.4} \\
  \searchrow
       & \modelhighlight
    & \textbf{4.6}\ci{1.9}
    & \textbf{29.8}\ci{6.5} & 18.9\ci{5.7} & \textbf{9.8}\ci{5.1}
    & 3.2\ci{3.4} & \textbf{26.3}\ci{4.4}
    & \textbf{20.6}\ci{3.5}\scoredelta{+1.4} \\
  \lightrule
  1.4B & \baselinehighlight
    & \textbf{8.0}\ci{2.4}
    & \textbf{44.7}\ci{7.0} & \textbf{22.8}\ci{6.1} & 8.3\ci{4.8}
    & \textbf{6.5}\ci{4.5} & 34.0\ci{4.8} & \textbf{27.2}\ci{3.9} \\
  \searchrow
       & \modelhighlight
    & 3.6\ci{1.7}
    & 41.5\ci{7.0} & 20.0\ci{5.8} & \textbf{12.9}\ci{5.7}
    & 2.4\ci{3.0} & 34.0\ci{4.8}
    & 26.2\ci{3.8}\scoredelta{-1.0} \\
  \lightrule
  3.3B & \baselinehighlight
    & 4.8\ci{1.9}
    & 39.9\ci{6.9} & 19.4\ci{5.8} & 8.3\ci{4.8} & \textbf{5.6}\ci{4.2} & 30.3\ci{4.6} & 24.2\ci{3.7} \\
  \searchrow
       & \modelhighlight
    & \textbf{5.6}\ci{2.0}
    & 39.9\ci{6.9} & \textbf{21.7}\ci{6.0} & \textbf{9.8}\ci{5.1}
    & 3.2\ci{3.4} & \textbf{32.7}\ci{4.7}
    & \textbf{25.4}\ci{3.8}\scoredelta{+1.2} \\
  \bottomrule
  \end{tabular}
  \end{adjustbox}
  \caption{\textbf{SFT model evaluation on \freshqa{}} by fact type and premise (LLM-judged accuracy, \%; 21 April 2026 snapshot, 500 test questions; zero-shot). Avg.: accuracy across all 500 questions. \nosearch{}: both models without search. 
  }
  \label{tab:freshqa}
  \end{table}

\halueval{} provides supported and hallucinated answer pairs; for \ragtruth{}, we construct analogous pairs from responses to the same question and retrieved evidence using the hallucination annotations.
We measure the fraction of examples in which the model assigns higher mean-token log probability to the supported answer than to the hallucinated answer (``prefer-right'') when no context, irrelevant context, and relevant context is provided. We place the contexts as plain text in the user message.

\begin{figure*}[t]
    \centering

    \begin{subfigure}[t]{0.49\linewidth}
      \centering
      \includegraphics[width=\linewidth]{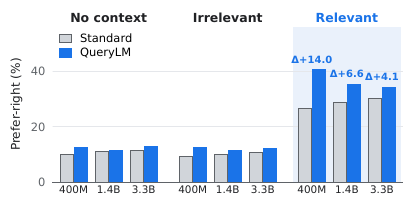}
      \caption{\halueval{} QA}
      \label{fig:grounding-halueval}
    \end{subfigure}\hfill
    \begin{subfigure}[t]{0.49\linewidth}
      \centering
      \includegraphics[width=\linewidth]{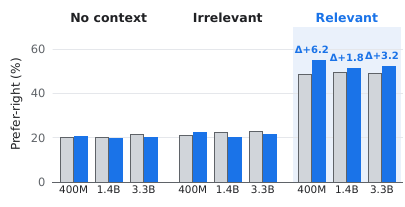}
      \caption{\ragtruth{}}
      \label{fig:grounding-ragtruth}
    \end{subfigure}

    \par\medskip

    \begin{subfigure}{\linewidth}
      \centering
      \small
      \setlength{\tabcolsep}{5pt}
      \begin{tabular}{l l c c c c c c}
        \toprule
        & & \multicolumn{3}{c}{\halueval{} QA}
          & \multicolumn{3}{c}{\ragtruth{}} \\
        \cmidrule(lr){3-5} \cmidrule(lr){6-8}
        Scale & Model
          & No context & Irrelevant & Relevant
          & No context & Irrelevant & Relevant \\
        \midrule
        400M & \baseline
          & 10.1\ci{0.6} & 9.2\ci{0.6} & 26.5\ci{0.9} & 20.1\ci{3.2} & 21.1\ci{3.3} & 48.6\ci{4.0} \\
        \searchrow
             & \modelhighlight
          & 12.4\ci{0.6} & 12.5\ci{0.6} & \textbf{40.5}\ci{1.0}
          & 20.6\ci{3.3} & 22.4\ci{3.3} & \textbf{54.8}\ci{4.0} \\
        \midrule
        1.4B & \baseline
          & 11.2\ci{0.6} & 10.1\ci{0.6} & 28.8\ci{0.9} & 20.1\ci{3.2} & 22.4\ci{3.3} & 49.6\ci{4.0} \\
        \searchrow
             & \modelhighlight
          & 11.3\ci{0.6} & 11.4\ci{0.6} & \textbf{35.4}\ci{0.9}
          & 19.7\ci{3.2} & 20.1\ci{3.2} & \textbf{51.4}\ci{4.0} \\
        \midrule
        3.3B & \baseline
          & 11.3\ci{0.6} & 10.9\ci{0.6} & 30.1\ci{0.9} & 21.6\ci{3.3} & 23.1\ci{3.4} & 49.2\ci{4.0} \\
        \searchrow
             & \modelhighlight
          & 12.9\ci{0.7} & 12.0\ci{0.6} & \textbf{34.2}\ci{0.9}
          & 20.1\ci{3.2} & 21.6\ci{3.3} & \textbf{52.4}\ci{4.0} \\
        \bottomrule
      \end{tabular}
      \caption{Prefer-right rates (\%).}
      \label{tab:groundedness-sft}
    \end{subfigure}

    \caption{\textbf{\halueval{} QA and \ragtruth{} evaluation}
    after \searchsft{}, prefer-right \%. a) and b) visualize the
    values in (c). 
    }
    \label{fig:grounding}
  \end{figure*}

\model{} favors supported statements on both \halueval{} and \ragtruth{} benchmarks (\Cref{tab:groundedness-sft}) with relevant context at every scale.
The gap in \model{} and baseline performance is smaller without context or with irrelevant context, suggesting that the lift in performance primarily comes from using relevant evidence to guide predictions.

\subsubsection{General Model Capability}
\label{app:general_capability}

Under a fixed training budget, \method{} allocates part of its compute to processing additional tokens introduced by search annotations.
To assess the effect on general capabilities, we evaluate \model{} and the FLOP-matched \baseline{} zero-shot on the held-out evaluation splits of eight benchmarks. For multiple-choice tasks, we rank answer options using conditional log-likelihood with length normalization and report per-task accuracy in \Cref{tab:general}.

\begin{table}[h]
\centering
\small
\setlength{\tabcolsep}{2.0pt}
\renewcommand{\arraystretch}{1.12}
\begin{adjustbox}{max width=\linewidth}
  \begin{tabular}{l l c c c c c c c c l}
\toprule
Scale & Model & ARC-E & ARC-C & HellaSwag & \piqa{} & WinoG. & OBQA & CSQA & LAMBADA & Avg. \\
\midrule
\grouprow{11}{Base models}
400M & \baselinehighlight & \textbf{56.7}\ci{2.0} & \textbf{30.7}\ci{2.6} & \textbf{45.7}\ci{1.0} & \textbf{69.3}\ci{2.1} & \textbf{52.9}\ci{2.7} & \textbf{34.0}\ci{4.1} & \textbf{19.8}\ci{2.2} & \textbf{41.4}\ci{1.3} & \textbf{43.8}\ci{0.9} \\
\searchrow
     & \modelhighlight & 53.2\ci{2.0} & 29.2\ci{2.6} & 43.3\ci{1.0} & 68.4\ci{2.1} & 52.2\ci{2.7} & 31.8\ci{4.1} & 18.8\ci{2.2} & 38.4\ci{1.3} & 41.9\ci{0.9}\scoredelta{-1.9} \\
\lightrule
1.4B & \baselinehighlight & 66.6\ci{1.9} & \textbf{38.7}\ci{2.8} & \textbf{60.2}\ci{1.0} & \textbf{74.3}\ci{2.0} & \textbf{59.7}\ci{2.7} & \textbf{39.2}\ci{4.3} & 18.9\ci{2.2} & \textbf{53.9}\ci{1.4} & \textbf{51.4}\ci{0.9} \\
\searchrow
     & \modelhighlight & \textbf{67.8}\ci{1.9} & 36.4\ci{2.8} & 58.0\ci{1.0} & 72.6\ci{2.0} & 59.4\ci{2.7} & 36.6\ci{4.2} & \textbf{19.7}\ci{2.2} & 51.7\ci{1.4} & 50.3\ci{0.9}\scoredelta{-1.1} \\
\lightrule
3.3B & \baselinehighlight & \textbf{73.8}\ci{1.8} & \textbf{44.2}\ci{2.8} & \textbf{65.5}\ci{0.9} & \textbf{76.6}\ci{1.9} & \textbf{65.0}\ci{2.6} & \textbf{42.8}\ci{4.3} & \textbf{21.0}\ci{2.3} & \textbf{59.0}\ci{1.3} & \textbf{56.0}\ci{0.9} \\
\searchrow
     & \modelhighlight & 70.2\ci{1.8} & 41.1\ci{2.8} & 62.2\ci{0.9} & 75.2\ci{2.0} & 62.6\ci{2.7} & 41.4\ci{4.3} & 19.7\ci{2.2} & 55.9\ci{1.4} & 53.5\ci{0.9}\scoredelta{-2.5} \\
     & \textcolor{searchblue}{$\model{}_{\text{Top-2}}$} & 75.2\ci{1.7} & 44.5\ci{2.8} & 65.8\ci{0.9} & 76.3\ci{1.9} & 64.0\ci{2.6} & 43.2\ci{4.3} & 19.5\ci{2.2} & 58.5\ci{1.3} & 55.9\ci{0.9}\scoredelta{-0.1} \\
\midrule
\grouprow{11}{After \searchsft{}}
400M & \baselinehighlight & \textbf{49.8}\ci{2.0} & \textbf{31.0}\ci{2.6} & \textbf{46.5}\ci{1.0} & \textbf{68.8}\ci{2.1} & 52.3\ci{2.7} & \textbf{36.4}\ci{4.2} & 19.7\ci{2.2} & \textbf{44.4}\ci{1.4} & \textbf{43.6}\ci{0.9} \\
\searchrow
     & \modelhighlight & 48.4\ci{2.0} & 30.6\ci{2.6} & 43.8\ci{1.0} & 67.6\ci{2.1} & \textbf{52.6}\ci{2.7} & 32.0\ci{4.1} & \textbf{21.9}\ci{2.3} & 41.0\ci{1.3} & 42.2\ci{0.9}\scoredelta{-1.4} \\
\lightrule
1.4B & \baselinehighlight & 52.5\ci{2.0} & \textbf{36.6}\ci{2.8} & \textbf{60.2}\ci{1.0} & \textbf{72.5}\ci{2.0} & 59.4\ci{2.7} & \textbf{41.8}\ci{4.3} & 18.3\ci{2.2} & 51.8\ci{1.4} & 49.1\ci{0.9} \\
\searchrow
     & \modelhighlight & \textbf{56.4}\ci{2.0} & 36.5\ci{2.8} & 58.9\ci{1.0} & 72.0\ci{2.1} & \textbf{59.8}\ci{2.7} & 40.6\ci{4.3} & \textbf{20.3}\ci{2.3} & \textbf{52.5}\ci{1.4} & \textbf{49.6}\ci{0.9}\scoredelta{+0.5} \\
\lightrule
3.3B & \baselinehighlight & 58.6\ci{2.0} & \textbf{41.7}\ci{2.8} & \textbf{65.1}\ci{0.9} & 74.3\ci{2.0} & \textbf{64.6}\ci{2.6} & \textbf{44.2}\ci{4.3} & \textbf{22.4}\ci{2.3} & \textbf{59.2}\ci{1.3} & \textbf{53.8}\ci{0.9} \\
\searchrow
     & \modelhighlight & \textbf{64.6}\ci{1.9} & 41.0\ci{2.8} & 62.7\ci{0.9} & \textbf{74.6}\ci{2.0} & 63.0\ci{2.7} & 43.4\ci{4.3} & 20.2\ci{2.3} & 55.4\ci{1.4} & 53.1\ci{0.9}\scoredelta{-0.7} \\
     & \textcolor{searchblue}{$\model{}_{\text{Top-2}}$} & 64.7\ci{1.9} & 42.8\ci{2.8} & 65.3\ci{0.9} & 75.8\ci{2.0} & 64.7\ci{2.6} & 43.0\ci{4.3} & 21.3\ci{2.3} & 56.1\ci{1.4} & 54.2\ci{0.9}\scoredelta{+0.4} \\
\midrule
\grouprow{11}{After standard SFT (Dolci mixture)}
400M & \baselinehighlight & \textbf{54.8}\ci{2.0} & \textbf{31.2}\ci{2.6} & \textbf{46.2}\ci{1.0} & \textbf{68.6}\ci{2.1} & \textbf{54.3}\ci{2.7} & \textbf{36.0}\ci{4.2} & 20.5\ci{2.3} & \textbf{46.1}\ci{1.4} & \textbf{44.7}\ci{0.9} \\
\searchrow
     & \modelhighlight & 52.1\ci{2.0} & 29.8\ci{2.6} & 44.1\ci{1.0} & 67.0\ci{2.1} & 53.5\ci{2.7} & 32.2\ci{4.1} & \textbf{21.2}\ci{2.3} & 43.1\ci{1.4} & 42.9\ci{0.9}\scoredelta{-1.8} \\
\lightrule
1.4B & \baselinehighlight & \textbf{65.8}\ci{1.9} & \textbf{40.3}\ci{2.8} & \textbf{60.7}\ci{1.0} & \textbf{74.1}\ci{2.0} & 60.6\ci{2.7} & 40.4\ci{4.3} & 19.2\ci{2.2} & \textbf{56.5}\ci{1.4} & \textbf{52.2}\ci{0.9} \\
\searchrow
     & \modelhighlight & 64.3\ci{1.9} & 37.0\ci{2.8} & 58.0\ci{1.0} & 72.9\ci{2.0} & \textbf{60.9}\ci{2.7} & \textbf{43.4}\ci{4.3} & \textbf{20.1}\ci{2.2} & 55.3\ci{1.4} & 51.5\ci{0.9}\scoredelta{-0.7} \\
\lightrule
3.3B & \baselinehighlight & \textbf{71.3}\ci{1.8} & \textbf{46.8}\ci{2.9} & \textbf{66.1}\ci{0.9} & \textbf{75.1}\ci{2.0} & \textbf{67.0}\ci{2.6} & \textbf{43.8}\ci{4.3} & \textbf{21.8}\ci{2.3} & \textbf{59.3}\ci{1.3} & \textbf{56.4}\ci{0.9} \\
\searchrow
     & \modelhighlight & 70.2\ci{1.8} & 42.7\ci{2.8} & 62.5\ci{0.9} & 74.7\ci{2.0} & 63.4\ci{2.6} & 42.4\ci{4.3} & 20.4\ci{2.3} & 57.3\ci{1.3} & 54.2\ci{0.9}\scoredelta{-2.2} \\
     & \textcolor{searchblue}{$\model{}_{\text{Top-2}}$} & 71.5\ci{1.8} & 44.6\ci{2.8} & 66.2\ci{0.9} & 75.8\ci{2.0} & 65.9\ci{2.6} & 44.0\ci{4.3} & 21.5\ci{2.3} & 60.2\ci{1.3} & 56.2\ci{0.9}\scoredelta{-0.2} \\
\bottomrule
\end{tabular}
  \end{adjustbox}
\caption{\textbf{General capabilities without search, \dclmedu{}} (accuracy, \%), for base models, after \searchsft{}, and after standard SFT (Dolci mixture). $\model{}_{\text{Top-2}}$: compute-matched 3.3B \model{} pretrained with the Top-2 augmentation (\Cref{sec:context_length_query_frequency}). Bold marks the better of \model{} and \baseline{}.
}
\label{tab:general}
\end{table}

\subsubsection{Switching Retrieval Source}
\label{app:retrieval_source_impact}

We evaluate the impact of switching the retrieval source on knowledge-intensive task accuracy.
Retrieval-enabled conditions use one search query per question, including for multi-hop tasks. \baseline{} models use \model{}'s queries and retrieved evidence; SFT models each generate their own queries. Tavily is restricted to Wikipedia for Wiki 360M models and searches the open web for DCLM models.
\Cref{tab:base-context-effect} shows the raw number used to plot \Cref{fig:context-ablation}, and \Cref{tab:sft-context-effect} shows additional evaluation result on SFT models.

\begin{figure*}[t]
    \centering
    \small
    \setlength{\tabcolsep}{2pt}
    \renewcommand{\arraystretch}{1.12}

    \begin{subfigure}[t]{\linewidth}
      \centering
      \begin{tabular*}{\linewidth}{@{\extracolsep{\fill}}l cc cc cc cc@{}}
      \toprule
      & \multicolumn{2}{c}{Wiki 360M}
      & \multicolumn{2}{c}{\shortstack{DCLM 400M\\(over-trained)}}
      & \multicolumn{2}{c}{DCLM 1.4B}
      & \multicolumn{2}{c}{DCLM 3.3B} \\
      \cmidrule(lr){2-3}\cmidrule(lr){4-5}
      \cmidrule(lr){6-7}\cmidrule(lr){8-9}
      Context
      & \baselinehighlight & \modelhighlight
      & \baselinehighlight & \modelhighlight
      & \baselinehighlight & \modelhighlight
      & \baselinehighlight & \modelhighlight \\
      \midrule
      Closed book
      & 3.7\ci{0.4} & 3.8\ci{0.4}
      & 11.5\ci{0.6} & 10.9\ci{0.6}
      & 15.2\ci{0.8} & 13.5\ci{0.7}
      & 16.7\ci{0.8} & 14.6\ci{0.8} \\
      \lightrule
      BM25 Wiki
      & 4.2\ci{0.3} & 5.7\ci{0.4}
      & 8.6\ci{0.5} & 17.0\ci{0.7}
      & 11.9\ci{0.6} & 20.4\ci{0.8}
      & 15.5\ci{0.7} & 22.0\ci{0.9} \\
      BM25 Sphere
      & \textbf{6.1}\ci{0.5} & \textbf{7.5}\ci{0.5}
      & 10.7\ci{0.6} & 20.5\ci{0.7}
      & 13.8\ci{0.7} & 23.8\ci{0.8}
      & 18.3\ci{0.8} & 25.5\ci{0.9} \\
      Tavily
      & 5.4\ci{0.4} & 7.0\ci{0.5}
      & \textbf{14.5}\ci{0.7} & \textbf{26.7}\ci{0.9}
      & \textbf{18.6}\ci{0.8} & \textbf{30.3}\ci{0.9}
      & \textbf{23.9}\ci{0.8} & \textbf{32.0}\ci{0.9} \\
      \bottomrule
      \end{tabular*}
      \caption{Base models (four-shot).}
      \label{tab:base-context-effect}
    \end{subfigure}

    \par\medskip

    \begin{subfigure}[t]{\linewidth}
      \centering
      \begin{tabular*}{\linewidth}{@{\extracolsep{\fill}}l cc cc cc cc@{}}
      \toprule
      & \multicolumn{2}{c}{Wiki 360M}
      & \multicolumn{2}{c}{\shortstack{DCLM 400M\\(over-trained)}}
      & \multicolumn{2}{c}{DCLM 1.4B}
      & \multicolumn{2}{c}{DCLM 3.3B} \\
      \cmidrule(lr){2-3}\cmidrule(lr){4-5}
      \cmidrule(lr){6-7}\cmidrule(lr){8-9}
      Context
      & \baselinehighlight & \modelhighlight
      & \baselinehighlight & \modelhighlight
      & \baselinehighlight & \modelhighlight
      & \baselinehighlight & \modelhighlight \\
      \midrule
      Closed book
      & 9.1\ci{0.6} & 7.0\ci{0.5}
      & 13.8\ci{0.6} & 8.9\ci{0.6}
      & 17.7\ci{0.7} & 8.1\ci{0.5}
      & 15.4\ci{0.7} & 5.0\ci{0.4} \\
      \lightrule
      BM25 Wiki
      & 7.9\ci{0.5} & 15.1\ci{0.6}
      & 16.6\ci{0.8} & 21.7\ci{0.9}
      & 25.1\ci{0.8} & 25.5\ci{0.9}
      & 28.4\ci{1.0} & 27.5\ci{0.9} \\
      BM25 Sphere
      & 8.4\ci{0.5} & 17.7\ci{0.7}
      & 17.4\ci{0.7} & 24.5\ci{0.8}
      & 27.6\ci{0.9} & 27.9\ci{0.9}
      & 30.6\ci{1.0} & 30.3\ci{0.9} \\
      Tavily
      & \textbf{9.5}\ci{0.5} & \textbf{21.4}\ci{0.9}
      & \textbf{25.4}\ci{0.9} & \textbf{30.2}\ci{0.9}
      & \textbf{33.6}\ci{1.0} & \textbf{35.4}\ci{1.0}
      & \textbf{36.8}\ci{1.1} & \textbf{36.9}\ci{1.0} \\
      \bottomrule
      \end{tabular*}
      \caption{SFT models (zero-shot).}
      \label{tab:sft-context-effect}
    \end{subfigure}

    \caption{\textbf{Effect of retrieval source on QA performance},
    Averaged answer-contained accuracy (\%) across all eight QA tasks. 
    }
    \label{fig:retrieval-source-effect}
  \end{figure*}

\subsubsection{Cost of \method{}}
\label{app:cost_method}

\Cref{tab:cost_table} summarizes the cost of \method{} in terms of added tokens and retrievals.

\begin{table}[t]
  \centering
  \small
  \setlength{\tabcolsep}{3pt}
  \renewcommand{\arraystretch}{1.12}
  \begin{tabular}{l c c c c c}
  \toprule
  & \multicolumn{2}{c}{Data preparation (once per corpus)}
  & Training
  & \multicolumn{2}{c}{Inference ($k$ lookups)} \\
  \cmidrule(lr){2-3} \cmidrule(lr){4-4} \cmidrule(lr){5-6}
  Method & Annotator tokens & Retrievals & Tokens & Tokens & Retrievals \\
  \midrule
  \baselinehighlight{} & -- & -- & $D$ & $T$ & -- \\
  \lightrule
  \searchrow \modelhighlight{} & $D + A\ell_a$ & $A$ & $D + A\ell$ & $T + k\ell$ & $k$ \\
  \bottomrule
  \end{tabular}
  \caption{\textbf{Cost of standard pretraining and \method{}} across data preparation, training and evaluation, in tokens processed and retrieval calls.
  $D$: original training corpus tokens; $A$: number of search queries;
  $\ell$: tokens added per search (search query, response, delimiters);
  $\ell_a$: annotator tokens computed per generated query;
  $T$: inference tokens;
  $k$: searches made at inference.}
  \label{tab:cost_table}
\end{table}

\end{document}